\documentclass[10pt,letterpaper,journal]{IEEEtran}
\usepackage{amsmath,amsfonts}
\usepackage{algorithm}
\usepackage{algorithmic}
\usepackage{array}
\usepackage[caption=false,font=normalsize,labelfont=sf,textfont=sf]{subfig}
\usepackage{textcomp}
\usepackage{stfloats}
\usepackage{url}
\usepackage{verbatim}
\usepackage{graphicx}
\usepackage{cite}
\usepackage{booktabs}
\usepackage{makecell}
\usepackage{multirow}
\usepackage[colorlinks,
            linkcolor=blue,
            anchorcolor=blue,
            citecolor=blue]{hyperref}
\usepackage{cuted}

\def\BibTeX{{\rm B\kern-.05em{\sc i\kern-.025em b}\kern-.08em
    T\kern-.1667em\lower.7ex\hbox{E}\kern-.125emX}}
\usepackage{balance}

\begin{document}

\title{Precise Time Delay Measurement and Compensation for Tightly Coupled Underwater SINS/piUSBL Navigation}

\author{Jin Huang,
\IEEEmembership{Graduate Student Member, IEEE},
Yingqiang Wang,
\IEEEmembership{Member, IEEE},
Haoda Li,
Zichen Liu,
Zhikun Wang,
Ying Chen
\thanks{Corresponding author: Yingqiang Wang and Ying Chen.}
\thanks{Jin Huang, Haoda Li, Zichen Liu, and Ying Chen are with the State Key Laboratory of Ocean Sensing, Ocean College of Zhejiang University, Zhoushan, 316021, China.}
\thanks{Yingqiang Wang is with the School of Oceanography, Shanghai Jiao Tong University, Shanghai, 200240, China. }
\thanks{Zhikun Wang is with the Donghai Laboratory, Zhoushan, 316021, China}
\thanks{\footnotesize © 2026 IEEE. Personal use of this material is permitted. Permission from IEEE must be obtained for all other uses, in any current or future media, including reprinting/republishing this material for advertising or promotional purposes, creating new collective works, for resale or redistribution to servers or lists, or reuse of any copyrighted component of this work in other works. DOI: 10.1109/TIM.2026.3676179\newline This work has been published in IEEE Transactions on Instrumentation and Measurement. The final published version is available at https://ieeexplore.ieee.org/document/11449348}
}


\maketitle

\noindent\textit{This is the author's accepted manuscript. The final published version is available at IEEE Xplore: https://ieeexplore.ieee.org/document/11449348}

\begin{abstract}
In multi-sensor systems, time synchronization is particularly challenging for underwater integrated navigation systems incorporating acoustic positioning, where time delays can significantly degrade accuracy when measurement and fusion epochs are misaligned.
This paper introduces a tightly coupled navigation framework that integrates a passive inverted ultra-short baseline (piUSBL) acoustic positioning system, a strapdown inertial navigation system (SINS), and a depth gauge under precise time synchronization.

The framework fuses piUSBL azimuth and slant range with depth measurements, avoiding poor vertical-angle observability in planar arrays.
By combining synchronized timing with acoustic signal processing, the proposed method transforms delay from an unobservable error into a measurable parameter, enabling explicit quantification of both acoustic propagation and system processing delays.
Field experiments demonstrate that the proposed approach reduces position RMSE by 44.02\% and maximum error by 40.79\% compared to the uncompensated baseline, while achieving further RMSE reductions of 37.66\% and 35.82\% in horizontal directions relative to filter-based delay compensation.
The results confirm that explicit delay measurement outperforms filter-based estimation, though instantaneous performance remains sensitive to acoustic signal quality, emphasizing the need for robust signal processing alongside accurate time synchronization in latency-sensitive multi-sensor systems.
\end{abstract}


\section{Introduction}
\IEEEPARstart{A}{utonomous} Underwater Vehicles (AUVs) are widely employed in various underwater applications, such as marine environment monitoring, underwater archaeology, and offshore structure inspection \cite{maStateArtKey2025, liImprovedESObasedLineofSight2025}. Due to the significant attenuation of electromagnetic signals in underwater environments, Global Navigation Satellite System (GNSS) signals cannot provide accurate positioning information for AUVs \cite{paullAUVNavigationLocalization2014}. In GNSS-denied environments, the strapdown inertial navigation system (SINS) is commonly used to provide continuous navigation information for AUVs \cite{zhangAutonomousUnderwaterVehicle2023}. However, the SINS is prone to error accumulation, leading to drift in the navigation solution over time \cite{chenReviewAUVUnderwater2015}. To address this issue, various aiding systems have been developed to assist the SINS and reduce drift errors, such as the Doppler velocity log (DVL), acoustic positioning systems, and altimeters \cite{millerAutonomousUnderwaterVehicle2010, duNovelLieGroup2022}.

Among various aiding systems, acoustic positioning system is widely used in underwater navigation due to its high accuracy and reliability.
The most common acoustic positioning systems are the long baseline (LBL), short baseline (SBL), and ultra-short baseline (USBL) systems \cite{maurelliAUVLocalisationReview2022}.
Of these, the USBL system is the most popular, primarily because of its compact size and ease of deployment on AUVs \cite{zhangNovelINSUSBL2022, zhaoReviewUnderwaterMultisource2023}.
To adapt to the expanding operational scenarios of AUVs, innovative USBL system derivations like inverted USBL (iUSBL) system and passive inverted USBL (piUSBL) system have been introduced \cite{vickeryAcousticPositioningSystems1998,rypkemaOnewayTraveltimeInverted2017}. 
In contrast to the conventional USBL system, the iUSBL architecture mounts the transducer array directly on the submersible, eliminating the need for position acquisition via acoustic communication with the surface vessel, which traditionally determines the relative position through two-way travel time (TWTT) measurements \cite{guoRobustAttitudeEstimation2023, yuUnderwaterLocalizationAUVs2023a}.
Building upon the iUSBL system, the piUSBL further introduces time synchronization and adopts a one-way travel time (OWTT) scheme for positioning \cite{rypkemaPassiveInvertedUltrashort2019,huangRaspi2USBLOpensourceRaspberry2025}.

Due to the relatively slow speed of sound propagation underwater, typically around 1500 m/s, the position information obtained from acoustic positioning systems often experiences delays ranging from several hundred milliseconds to a few seconds \cite{zhaoUnsynchronizedUnderwaterLocalization2024}. 
For an integrated navigation system (INS) using inertial measurement unit (IMU) with sampling rates exceeding 200 Hz, these delays can accumulate significant errors and compromise system accuracy if not properly addressed \cite{ridaoUSBLDVLNavigation2011,fossenFeedbackErrorstateKalman2023}.

Research on this issue can be broadly categorized into two directions: estimating the delay and compensating for the delay. 
The first direction focuses on estimating or eliminating the observed time delay. 
The method typically augments the state vector with a delay variable, estimates the delay using slant range and vehicle velocity information, and then applies compensation directly \cite{xiaMixtureDistributionBasedRobust2021, wangAdaptiveUnscentedKalman2025a}.
In addition, inspired by GNSS, a dual-beacon/hydrophone differential method eliminates the common errors by differencing the measurements from two beacons/hydrophones, which includes the clock measurement bias, transponder delay, and sound speed error\cite{liuRobustTightlySINS2023a, xuRobustIterativeAlgorithm2025, zhangNovelUnderwaterSINS2025}. 
Despite their effectiveness, these methods do not analyze the observability of the time delay in the error state model, and the accuracy of time estimation remains uncertain. 
In differential approaches, the inter-system delay and observability issues also require further study.

The second direction assumes that the delay is already known and focuses on designing filters that can properly process delayed sensor data. 
Existing strategies can be classified into two main categories.
The first is retroactive processing, which revises past states using techniques such as filter recalculation (FR) \cite{comelliniIncorporatingDelayedMultirate2020}, state inversion (SI) \cite{xuMaximumCorrentropyDelay2021a}, or bounded circular buffers (BCB) to store historical states \cite{ribasDelayedStateInformation2012}. 
The second is real-time processing, where delayed observations are handled as they arrive. 
This includes approaches such as velocity aiding with DVL measurements, projecting delayed data onto the current state with adjusted measurement noise \cite{xiaoAcousticCommunicationTime2017}, or reconstructing historical states through state transition matrix inversion to form equivalent measurement updates \cite{xuNovelRobustFilter2021a}.
Another representative method is the delay-state unscented Kalman filter (DS-UKF), which incorporates historical states into an extended state vector and processes them once the delayed measurements become available \cite{wangInformationFusionAlgorithm2023a}.

Overall, existing approaches predominantly rely on estimating sensor delays and then compensating for them within the filtering framework.
However, most of these methods treat the delay as a nuisance parameter to be estimated, while the estimation accuracy and, more fundamentally, the observability of the delay within the error-state model are often not explicitly analyzed.
As a result, delay compensation may become fragile when the delay is time-varying or weakly observable.

To address these limitations, this paper introduces a tightly coupled architecture that leverages the piUSBL system as an acoustic sensing front-end with synchronized timing support.
Unlike conventional acoustic positioning pipelines, where the effective measurement epoch is ambiguous, the piUSBL provides a practical basis for assigning accurate timestamps to the key stages of signal acquisition and processing via a synchronous clock mechanism \cite{rypkemaUnderwaterOutSight2019}. 
Importantly, the contribution of this work goes beyond using synchronized clocks as a hardware upgrade by exploiting the synchronized timing to explicitly measure the end-to-end delay and to integrate the resulting effective measurement epoch into the navigation architecture, enabling online compensation of both acoustic propagation and processing delays throughout the fusion process.

Specifically, this paper proposes a tightly coupled SINS/piUSBL/depth-gauge integrated navigation system with precise time synchronization. 
The framework fuses the piUSBL azimuth and slant-range measurements with the depth-gauge observation, which provides complementary constraints on the vertical channel and introduces attitude sensitivity through the depth-gauge innovation Jacobian, thereby mitigating the lack of vertical-angle information inherent to planar arrays.
By utilizing synchronized timing to measure the delay, the proposed framework avoids relying solely on delay estimation and instead turns the delay from a weakly observable error source into a measurable quantity that can be explicitly compensated in the filter.

The main contributions of this paper are summarized as follows:

\begin{itemize}
    \item [1.] \textbf{A tightly coupled SINS/piUSBL/depth-gauge navigation architecture.} This paper formulates a compact measurement model that jointly uses piUSBL azimuth and slant range with depth as complementary observations, and analyzes the local weak observability of navigation-relevant states under nominal sensor availability, thereby mitigating the lack of vertical-angle information inherent to planar arrays.
    \item [2.] \textbf{A delay measurement and compensation mechanism enabled by synchronized timing.} By integrating time synchronization into the acoustic processing chain and the fusion architecture, we quantify the effective measurement epoch and compensate for both acoustic propagation and system processing delays. This transforms the delay from a latent error term into a measurable parameter that can be explicitly handled in the tightly coupled filter.
    \item [3.] \textbf{Quantitative evaluation under simulation and field conditions.} Simulations analyze the impact of time delays on acoustic positioning performance and demonstrate the benefit of the proposed delay compensation strategy, while field experiments further validate the proposed tightly coupled system under real measurement conditions. Together, these evaluations highlight not only empirical improvements but also the general applicability of the approach to delay-sensitive navigation scenarios.
\end{itemize}

The rest of this paper is organized as follows: Section \ref{sec:system_architecture} introduces the architecture of the proposed tightly coupled integrated navigation system with time synchronization. Section \ref{sec:Methodology} presents the methodology, including the integrated navigation strategy utilizing precise time delay measurements, the SINS/piUSBL/depth gauge tightly coupled integrated navigation algorithm, and the observability analysis. Section \ref{sec:simulation_and_field_experiments} describes the simulation and field experiments conducted to validate the proposed approach. Finally, Section \ref{sec:conclusion} concludes the paper.

\section{System Architecture}
\label{sec:system_architecture}

In most acoustic-aided integrated navigation systems, the acoustic subsystem operates independently and is loosely coupled with the overall system, which limits researchers’ access to its raw data, such as per-channel signals and detailed processing procedures.
In this paper, we propose a navigation framework that deeply integrates acoustic positioning with inertial navigation, realized through a tightly coupled SINS/piUSBL/depth gauge system with time synchronization.
For simplicity, the term “SINS/piUSBL” is hereafter used to denote the SINS/piUSBL/depth gauge system.

\begin{figure}[!t]
    \centering 
    \includegraphics[width=\linewidth]{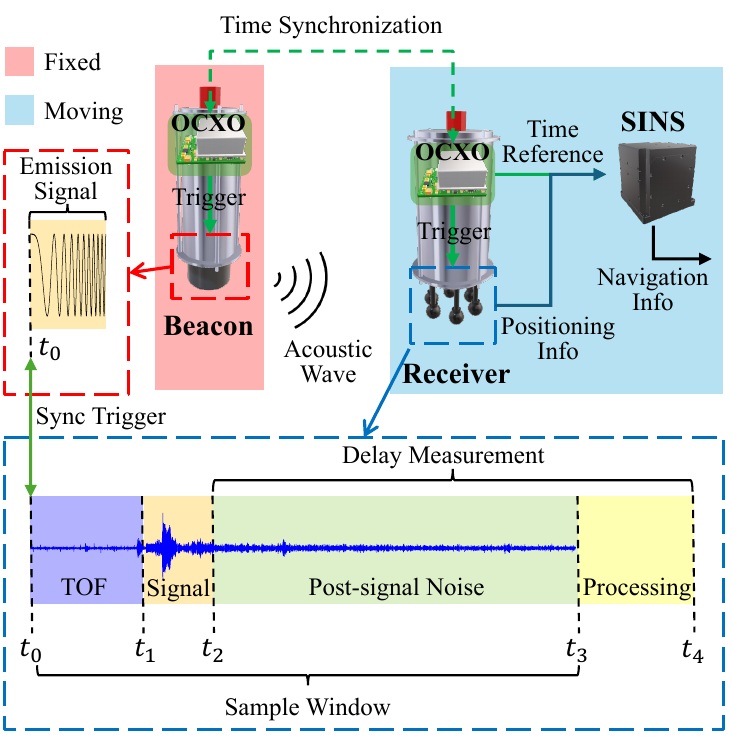} 
    \caption{The system architecture of the SINS/piUSBL tightly coupled navigation system with time synchronization.} 
    \label{fig:SystemInfo}
\end{figure}

Fig. \ref{fig:SystemInfo} illustrates the essential components of the SINS/ piUSBL integrated navigation system. The beacon is mounted on a static platform, such as a surface vessel, while the receiver is installed on a submersible that is integrated with a SINS unit.
Both the beacon and the receiver are equipped with synchronized clocks, such as the oven-controlled crystal oscillator (OCXO) used in this study, which cost approximately 100 USD.
Prior to mission execution, the clocks on both sides are disciplined using a high-precision GNSS time source to ensure precise time alignment and synchronization.

The OCXO generates a one pulse-per-second (1 PPS) signal, which is used to trigger the beacon to transmit an acoustic signal at the beginning of each second and the receiver synchronizes its sampling process with this signal.
A matched filter is then applied by the onboard processor to detect the signal's time of flight (TOF), followed by conventional beamforming to estimate the azimuth of the beacon. 
After signal processing, the slant range and azimuth are transmitted to the SINS system for measurement updates. 
In time $t_0$, the OCXO triggers the beacon to transmit an acoustic signal and simultaneously initiates the receiver's sampling process, respectively. Time $t_1$ marks the moment when the receiver detects the begining of the incoming signal, while time $t_2$ indicates the end of the designed signal length. The sampling continues until time $t_3$, which marks the end of the sampling window. Finally, the processed results are delivered to the SINS system at time $t_4$.
In the time $t_1$, the paper defines that the TOF and direction of arrival (DOA) can be determined, even if the emission signal is not be fully record.
Thus, the delay measured can be defined as the period between $t
_1$ to $t_4$.

\begin{figure}[!t]
    \centering
    \includegraphics[width=\linewidth]{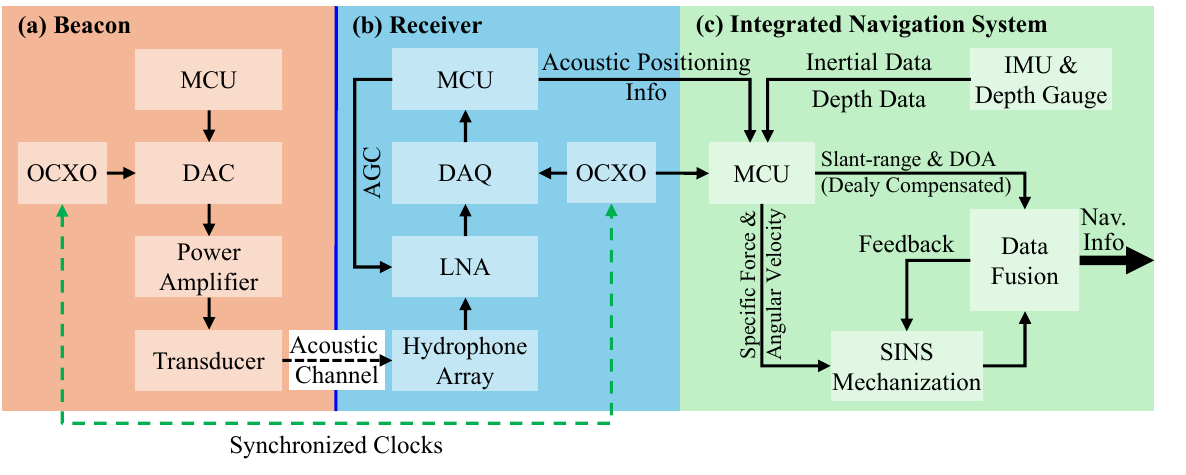}
    \caption{The diagram of the SINS/piUSBL/depth gauge tightly coupled navigation system with time synchronization.}
    \label{fig:SystemFlowChart}
\end{figure}

\begin{figure}[!t]
    \centering 
    \includegraphics[width=\linewidth]{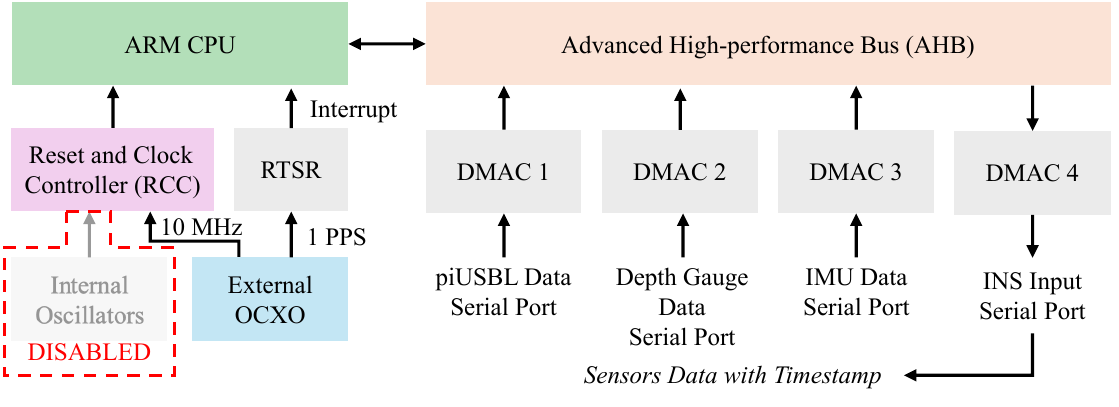} 
    \caption{The architecture of the microcontroller unit (MCU) with the OCXO for time synchronization.} 
    \label{fig:STM32FlowChart}
\end{figure}

The architecture of the piUSBL acoustic positioning system with OCXO, detailed in \cite{wangDesignExperimentalResults2022, wangPassiveInvertedUltraShort2022}, along with its integration with the SINS, is illustrated in Fig. \ref{fig:SystemFlowChart}.
In the integrated navigation system, an STM32 microcontroller unit (MCU) is responsible for managing timestamps across multiple sensors, including a fiber-optic gyroscope (FOG)-based IMU, a depth gauge, and the piUSBL acoustic positioning system.

The OCXO from the piUSBL system provides a highly stable clock reference for the integrated navigation system, particularly serving as the primary timing source for the MCU, as illustrated in Fig. \ref{fig:STM32FlowChart}.
For the MCU, its internal oscillator is replaced by the external 10 MHz square-wave signal generated by the OCXO, which functions as the high-speed external oscillator (HSE) for system clock generation.
Furthermore, the OCXO outputs a 1 PPS signal that triggers both the piUSBL system’s transmission and sampling processes, and the MCU’s rising-edge trigger selection register (RTSR) to generate an external interrupt.
Upon receiving the 1 PPS signal, the central processing unit (CPU) records the corresponding timestamp, which corresponds to the time of sampling start within the piUSBL system.
Meanwhile, the advanced high-performance bus (AHB) receives sensor data via the direct memory access controller (DMAC), including measurements from the piUSBL, depth gauge, and IMU, and transfers them to the MCU for timestamp alignment and subsequent output.
Through clock synchronization between the piUSBL system and the SINS, the timestamps of the acoustic measurements can be precisely aligned with external sensor data, thereby enabling accurate time-delay measurement and compensation.

\section{Methodology}
\label{sec:Methodology}

The integration of synchronized clocks enables the system to record accurate timing information across various components.
This section describes how these timestamps are used to calculate and handle the acoustic delay, and how the piUSBL observations are incorporated into the SINS for measurement updates.

\subsection{Precise Time Delay Measurements}
\label{sec:integrated_navigation_strategy_utilizing_precise_time_delay_measurements}

In practice, a significant time delay exists between the reception of the acoustic signal and the availability of the computed position from the piUSBL system.

As mentioned in Fig. \ref{fig:SystemInfo}, the interval $t_2 - t_1$ corresponds to the designed signal length, typically 20 ms in this study.
Sampling continues until time $t_3$, marking the end of the sampling window. 
The duration of this window ($t_3 - t_0$) is determined by the maximum expected range of the piUSBL system, which is typically set to 2 seconds during sea trials.
The onboard processor applies matched filtering and beamforming to the sampled data to estimate the slant range and azimuth to the beacon. 
The processed results are then delivered to the SINS system at time $t_4$.

\begin{figure}[!t]
\centering
\includegraphics[width=\linewidth]{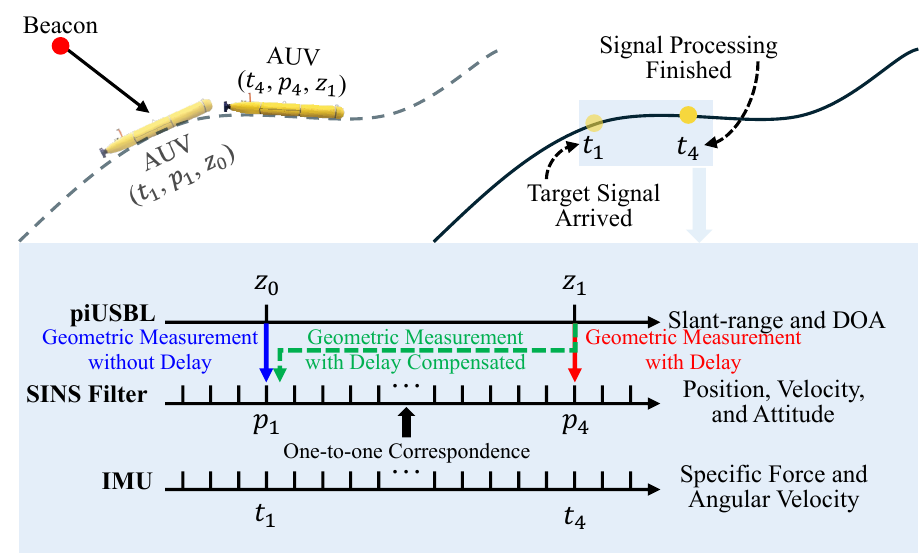}
\caption{The delay illustration of the SINS/piUSBL tightly coupled navigation system.}
\label{fig:DelayIllustrate}
\end{figure}

As shown in Fig. \ref{fig:DelayIllustrate}, at time $t_1$, the AUV is located at position $p_1$, which corresponds to the ideal position associated with the piUSBL measurement $z_0$. However, the slant range and azimuth are not available until time $t_4$, when the processed measurement $z_1$ corresponds to the AUV's position $p_4$.
In this paper, the delay of the piUSBL system is defined as the time difference between the actual measurement time and the time when the measurement becomes available, which is expressed as:

\begin{equation}
\label{eq:delay_definition}
\Delta t = t_4 - t_1
\end{equation}

In practical applications, due to time synchronization, the true timestamp of the piUSBL measurement used in the SINS filter can be computed as:

\begin{equation}
\label{eq:real_timestamp}
t_1 = t_0 + t_{\text{tof}}
\end{equation}

Here, $t_0$ is the time when the PPS signal is received and recorded by the MCU, and $t_{\text{tof}}$ is the time of flight (TOF) of the acoustic signal, which can be obtained from the matched filtering process.

Once the true timestamp $t_1$ is accurately determined, the filtering reconstruction method is employed to address the delay issue within the navigation filter.

\subsection{SINS/piUSBL Tightly Coupled Navigation}
\label{sec:sinspisusbldepth_tightly_coupled_navigation}

The involved reference frames in the following sections are the earth-centered-earth-fixed (ECEF) frame (denoted as $e$-frame), the navigation frame (denoted as $n$-frame), the body frame (denoted as $b$-frame), the depth gauge frame (denoted as $d$-frame), and the piUSBL frame (denoted as $u$-frame).
These frames follow the community convention which can be found in \cite{niuKFGINSOpensourcedSoftware2025}.

\begin{figure}[!t]
    \centering 
    \includegraphics[width=\linewidth]{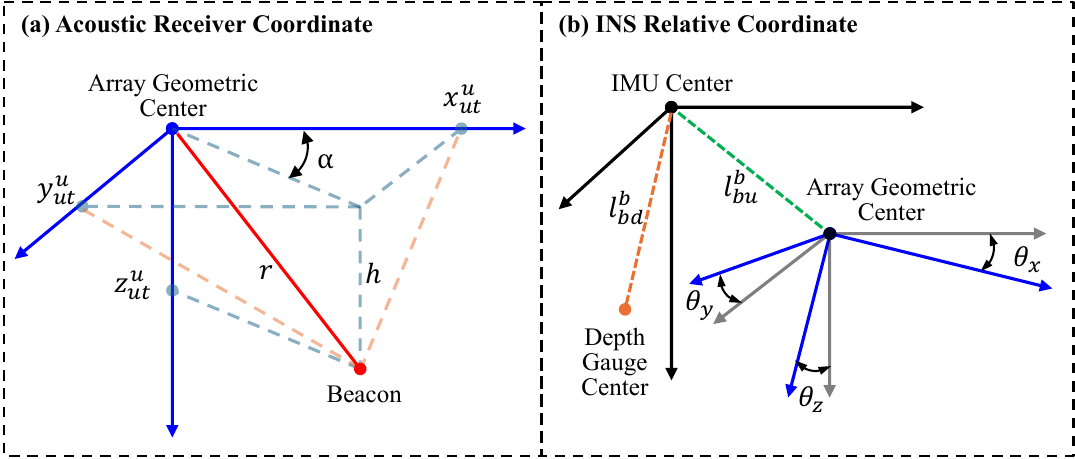} 
    \caption{The Coordinate System of the tightly coupled navigation system.} 
    \label{fig:Coordinate} 
\end{figure}

Fig. \ref{fig:Coordinate} illustrates the observation model of the piUSBL system and the relative positions of the IMU, piUSBL, and depth gauge.
In Fig. \ref{fig:Coordinate} (a), the $u$-frame is defined with its origin at the hydrophone phase center of the piUSBL system, the $x$-axis pointing forward along the array's first hydrophone, the $y$-axis pointing to the right, and the $z$-axis pointing downward.
And the piUSBL provides two types of observations in the $u$-frame: the slant range $r$ and the azimuth $\alpha$ of the beacon.
It is worth noting that, compared with the conventional USBL system described in \cite{xiaMixtureDistributionBasedRobust2024}, which provides two angular measurements and one slant range, the piUSBL system employed in this study lacks one angular measurement due to its planar circular array configuration.
However, a single azimuth combined with the slant range is insufficient to fully determine the 3-dimensional position of the beacon in the $u$-frame. 
To address this limitation, a depth gauge is integrated into the tightly coupled navigation system to provide depth information for the submersible.
As a compact and portable acoustic positioning solution, piUSBL has a limited effective range. Therefore, it is reasonable to assume that the horizontal plane remains approximately flat within the working area. 
Under the assumption that the beacon's position is known, and by combining the $u$-frame attitude, the vertical distance $h$ in the $u$-frame can be readily calculated.
The coordinates, $\boldsymbol{P}_{ut}^u = \left[ x_{ut}^u, y_{ut}^u, z_{ut}^u \right]^\top$, represent the projected 3-dimensional position of the beacon in the $u$-frame and are computed as follows:

\begin{equation}
    \label{eq:piUSBL_observation}
    \left\{
    \begin{aligned}
        x_{ut}^u &= \sqrt{r^2 - h^2} \cos{\alpha} \\
        y_{ut}^u &= \sqrt{r^2 - h^2} \sin{\alpha} \\
        z_{ut}^u &= h
    \end{aligned}
    \right.
\end{equation}
where $h$ can be calculated as follows:
\begin{equation}
    \label{eq:piUSBL_depth}
    h = \mathbf{C}_b^u \mathbf{C}_n^b \mathbf{C}_e^n (h_t^e - h_u^e) 
\end{equation}
where the $\mathbf{C}_b^u$, $\mathbf{C}_n^b$, and $\mathbf{C}_e^n$ are the direction cosine matrices (DCMs) representing the transformation from the $b$-frame to the $u$-frame, from the $n$-frame to the $b$-frame, and from the $e$-frame to the $n$-frame, respectively.
The parameters $h_t^e$ and $h_u^e$ refer to the altitudes of the beacon and the hydrophone phase center of the piUSBL system in the $e$-frame, respectively.

As shown in Fig. \ref{fig:Coordinate}(b), $\boldsymbol{l}_{bu}^b$ and $\boldsymbol{l}_{bd}^b$ represent the lever arms from the IMU to the piUSBL and the depth gauge, respectively, and both are expressed in the body frame.
The piUSBL installation misalignment is modeled by the error vector $\delta \boldsymbol{\theta} = \left[ \theta_x, \theta_y, \theta_z \right]^\top$. The lever arms and installation misalignment are determined through precise calibration before deployment \cite{wangAdaptiveUnscentedKalman2025,huangGNSSaidedInstallationError2025}. 

Considering the aforementioned sources of error and the typical 21-dimensional error state equation of the INS \cite{niuDevelopmentEvaluationGNSS2015}, the error state of the tightly coupled navigation system can be expanded as follows:

\begin{equation}
    \label{eq:error_state}
    \boldsymbol{X} = \left[ \delta \boldsymbol{p}^n ; \delta \boldsymbol{v}^n; \boldsymbol{\phi}; \boldsymbol{g}_b; \boldsymbol{a}_b; \boldsymbol{g}_s; \boldsymbol{a}_s; \delta \boldsymbol{\theta}; \delta k; \delta h^n \right]^\top_{(26 \times 1)}
\end{equation}
where $\delta \boldsymbol{p}^n$, $\delta \boldsymbol{v}^n$, and $\boldsymbol{\phi}$ denote the position, velocity, and attitude errors in the navigation frame ($n$-frame), respectively.
$\boldsymbol{g}_b$ and $\boldsymbol{a}_b$ represent the bias errors of the gyroscope and accelerometer, while $\boldsymbol{g}_s$ and $\boldsymbol{a}_s$ denote their corresponding scale factor errors \cite{grovesPrinciplesGNSSInertial2013}.
$\delta \boldsymbol{\theta}$ denotes the installation misalignment error of the piUSBL system, $\delta k$ is the scale factor error associated with the slant range measurement of the piUSBL, and $\delta h^n$ represents the depth gauge measurement error in the $n$-frame .

The state transition matrix $\boldsymbol{F}$ of the tightly coupled navigation system can be expressed as follows:

\begin{equation}
    \label{eq:state_transition_matrix}
    \boldsymbol{F} = 
    \begin{bmatrix}
        \boldsymbol{F}_{\text{SINS}} & \boldsymbol{0}_{21\times 4} & \boldsymbol{0}_{21\times 1} \\
        \boldsymbol{0}_{4\times 21} & \boldsymbol{F}_{\text{piUSBL}} & \boldsymbol{0}_{4\times 1} \\
        \boldsymbol{0}_{1\times 21} & \boldsymbol{0}_{1\times 4} & F_{\text{Depth}} \\
    \end{bmatrix}_{(26 \times 26)}
\end{equation}
where the SINS state transition matrix, $\boldsymbol{F}_{\text{SINS}}$, is described in \cite{tittertonStrapdownInertialNavigation2004}.
The installation misalignment error $\delta \boldsymbol{\theta}$, the slant range scale factor error $\delta k$, and the depth measurement error $\delta h^n$ are assumed to be constant or slowly varying, with negligible dynamics. 
Therefore, their corresponding state transition matrices are modeled as zero: $\boldsymbol{F}_{\text{piUSBL}} = \boldsymbol{0}_{4\times 4}$, $F_{\text{Depth}} = 0$.

In the construction of measurement equations, the piUSBL system provides two types of measurements: the slant range $r$ and the azimuth $\alpha$, while the depth gauge provides the vertical distance $h^n$. 
The measurement equations can be expressed as follows:

\begin{equation}
    \label{eq:measurement_equation}
    \boldsymbol{Z} = 
    \begin{bmatrix}
        \boldsymbol{Z}_{\text{piUSBL}} \\
        Z_{\text{Depth}}
    \end{bmatrix} = 
    \begin{bmatrix}
        \hat{r}_{\text{SINS}} - \tilde{r}_{\text{piUSBL}} \\
        \hat{\alpha}_{\text{SINS}} - \tilde{\alpha}_{\text{piUSBL}} \\
        \hat{h}_{\text{SINS}}^n - \tilde{h}_{\text{Depth}}^n \\
    \end{bmatrix}
    = \boldsymbol{H} \boldsymbol{X} + \boldsymbol{V}
\end{equation}
where $\boldsymbol{Z}$ is the measurement vector, $\hat{r}_{\text{SINS}}$, $\hat{\alpha}_{\text{SINS}}$, and $\hat{h}_{\text{SINS}}^n$ are the estimated slant range, azimuth, and depth, respectively.
The tilde variables $\tilde{r}_{\text{piUSBL}}$, $\tilde{\alpha}_{\text{piUSBL}}$, and $\tilde{h}_{\text{Depth}}^n$ represent the actual measurements from the piUSBL system and the depth gauge, respectively.
$\boldsymbol{H}$ is the measurement matrix, and $\boldsymbol{V}$ is the measurement noise vector.

Then, the measurement matrix $\boldsymbol{H}$ in the tightly coupled navigation system is derived.
Considering \eqref{eq:piUSBL_observation}, the relationship between the coordinates of the beacon in the $u$-frame and the output of the piUSBL system is expressed as follows:

\begin{equation}
    \label{eq:piUSBL_output}
    \begin{bmatrix}
        r \\
        \alpha \\
    \end{bmatrix} = 
    \begin{bmatrix}
        \sqrt{{x_{ut}^u}^2 + {y_{ut}^u}^2 + {z_{ut}^u}^2} \\
        \arctan2 \left({y_{ut}^u},{x_{ut}^u}\right) \\
    \end{bmatrix}
\end{equation}

Considering the lever arm from the IMU to the piUSBL, the coordinates of the beacon in the $u$-frame are expressed in terms of its coordinates in the $b$-frame as follows:

\begin{equation}
    \label{eq:putu_from_putb}
    \boldsymbol{P}_{ut}^u = \mathbf{C}_b^u \left( \boldsymbol{P}_{bt}^b - \boldsymbol{l}_{bu}^b \right)
\end{equation}

The Taylor expansion of \eqref{eq:piUSBL_output} around the estimated position $\boldsymbol{P}_{ut}^u$ is expressed as follows:

\begin{equation}
    \label{eq:delta_r_alpha}
    \begin{bmatrix}
        \delta r & \delta \alpha \\
    \end{bmatrix}^\top =
    \boldsymbol{H}_a 
    \begin{bmatrix}
        \delta x_{ut}^u &
        \delta y_{ut}^u &
        \delta z_{ut}^u \\
    \end{bmatrix}^\top
    = \boldsymbol{H}_a \delta \boldsymbol{P}_{ut}^u
\end{equation}
where $\boldsymbol{H}_a$ is the Jacobian matrix, which can be expressed as follows:

\begin{equation}
    \label{eq:Ha}
    \boldsymbol{H}_a = 
    \begin{bmatrix}
        \frac{x_{ut}^u}{r} & \frac{y_{ut}^u}{r} & \frac{z_{ut}^u}{r} \\
         - \frac{y_{ut}^u}{{x_{ut}^u}^2 + {y_{ut}^u}^2} & \frac{x_{ut}^u}{{x_{ut}^u}^2 + {y_{ut}^u}^2} & 0 \\
    \end{bmatrix}
\end{equation}

According to \eqref{eq:putu_from_putb}, the partial derivative of $\boldsymbol{P}_{ut}^u$ with respect to $\boldsymbol{P}_{bt}^b$ can be given by:

\begin{equation}
    \label{eq:partial_putu}
    \frac{\partial \boldsymbol{P}_{ut}^u}{\partial \boldsymbol{P}_{bt}^b} = \mathbf{C}_b^u
\end{equation}

The error in $\boldsymbol{P}_{ut}^u$ caused by the SINS error is analyzed. 
The coordinates of the beacon in the $u$-frame are expressed in the vector in the $e$-frame as follows:

\begin{equation}
    \label{eq:putu_from_pute}
    \boldsymbol{P}_{ut}^u = \mathbf{C}_b^u \mathbf{C}_n^b \mathbf{C}_e^n \boldsymbol{P}_{ut}^e
\end{equation}
where $\boldsymbol{P}_{ut}^e$ represents the coordinates of the beacon in the $e$-frame, which are expressed as follows:

\begin{equation}
    \label{eq:pute}
    \begin{aligned}
        \boldsymbol{P}_{ut}^e 
        & = \boldsymbol{r}_{t}^e - \boldsymbol{r}_{u}^e \\
        & = \boldsymbol{r}_{t}^e - \boldsymbol{r}_{b}^e - \mathbf{C}_n^e \mathbf{C}_b^n  \boldsymbol{l}_{bu}^b \\
        & = \boldsymbol{P}_{bt}^e - \mathbf{C}_n^e \mathbf{C}_b^n  \boldsymbol{l}_{bu}^b \\
    \end{aligned}
\end{equation}
where $\boldsymbol{P}_{bt}^e$ is the relative position from the body frame origin to the beacon in the $e$-frame, $\boldsymbol{r}_{t}^e$ is the absolute position of the beacon in the $e$-frame, $\boldsymbol{r}_{u}^e$ is the absolute position of the hydrophone phase center of the piUSBL system in the $e$-frame, and $\boldsymbol{l}_{bu}^b$ is the lever arm from the IMU to the piUSBL in the $b$-frame.

By applying a perturbation to \eqref{eq:putu_from_pute} and substituting \eqref{eq:pute} into it, the corresponding error expression is derived as:

\begin{equation}
    \label{eq_putu_perturbation}
    \begin{aligned}
        \hat{\boldsymbol{P}}_{ut}^u 
        & = \hat{\mathbf{C}}_b^u \hat{\mathbf{C}}_n^b \hat{\mathbf{C}}_e^n \hat{\boldsymbol{P}}_{ut}^e \\
        & = \hat{\mathbf{C}}_b^u \hat{\mathbf{C}}_n^b \hat{\mathbf{C}}_e^n \hat{\boldsymbol{P}}_{bt}^e - \hat{\mathbf{C}}_b^u \boldsymbol{l}_{bu}^b
    \end{aligned}
\end{equation}

We now proceed to expand each term individually as \eqref{eq:hat_cen}.

\begin{figure*}[!h]
\begin{equation}
\label{eq:hat_cen}
\begin{aligned}
    \hat{\mathbf{C}}_e^n
    & = \mathbf{C}_e^n + \delta \mathbf{C}_e^n \\
    & = 
    \begin{bmatrix}
        \sin \left( L+\delta L \right) \cos \left( \lambda +\delta \lambda \right)&		-\sin \left( L+\delta L \right) \sin \left( \lambda +\delta \lambda \right)&		\cos \left( L+\delta L \right)\\
        -\sin \left( \lambda +\delta \lambda \right)&		\cos \left( \lambda +\delta \lambda \right)&		0\\
        -\cos \left( L+\delta L \right) \cos \left( \lambda +\delta \lambda \right)&		-\cos \left( L+\delta L \right) \sin \left( \lambda +\delta \lambda \right)&		-\sin \left( L+\delta L \right)\\
    \end{bmatrix} \\
\end{aligned}
\end{equation}

\begin{equation}
\label{eq:delta_cen}
\delta \mathbf{C}_{e}^{n} \approx
    \begin{bmatrix}
	-\delta L\cos L \cos \lambda+\delta \lambda \sin \lambda \sin L&		-\delta \lambda \cos \lambda \sin L-\delta L\cos L\sin \lambda&		-\delta L\sin L\\
	-\delta \lambda \cos \lambda&		-\delta \lambda \sin \lambda&		0\\
	\delta L\sin L\cos \lambda +\delta \lambda \sin \lambda \cos L&		\delta L\sin L\sin \lambda -\delta \lambda \cos \lambda \cos L&		-\delta L\cos L\\
    \end{bmatrix}
\end{equation}
\end{figure*}

Simplifying the second-order error term in \eqref{eq:hat_cen}, the error of the DCM from the $e$-frame to the $n$-frame can be expressed as \eqref{eq:delta_cen}, where $\delta L$ and $\delta \lambda$ are the SINS latitude and longitude errors, respectively.

Under the small-angle approximation, the perturbed DCM from the $b$-frame to the $n$-frame in the SINS is represented as:

\begin{equation}
\label{eq:hat_cnb}
\hat{\mathbf{C}}_n^b = \mathbf{C}_n^b \left(\boldsymbol{I} + \boldsymbol{\phi} \times \right)
\end{equation}
where $\boldsymbol{\phi} \times$ is the skew-symmetric matrix of the attitude error vector $\boldsymbol{\phi}$.

Since the beacon's position $\boldsymbol{r}_t^e$ and the lever arm $\boldsymbol{l}_{bu}^b$ can be accurately measured, $\hat{\boldsymbol{P}}_{bt}^e$ can be expressed as:

\begin{equation}
\label{eq:rte}
\boldsymbol{r}_t^e =
\begin{bmatrix}
    \left( R_N+h_t \right) \cos L_{t}\cos \lambda _t\\
	\left( R_N+h_t \right) \cos L_{t}\sin \lambda _t\\
	\left( R_M+h_t \right) \sin L_{t}
\end{bmatrix} 
\end{equation}

\begin{equation}
\label{eq:hat_rbe}
\hat{\boldsymbol{r}}_b^e =
\begin{bmatrix}
    \left( R_N+\hat{h}_b \right) \cos \hat{L}_{b}\cos \hat{\lambda}_b\\
	\left( R_N+\hat{h}_b \right) \cos \hat{L}_{b}\sin \hat{\lambda}_b\\
	\left( R_M+\hat{h}_b \right) \sin \hat{L}_{b}
\end{bmatrix} 
\end{equation}

\begin{equation}
    \label{eq:hat_pbte}
    \hat{\boldsymbol{P}}_{bt}^e = \boldsymbol{r}_t^e - \hat{\boldsymbol{r}}_b^e
\end{equation}
where $R_N$ and $R_M$ denote the radius of curvature of the earth in the prime vertical and meridian directions, respectively, $h_t$ and $h_b$ represent the altitudes of the beacon and the IMU in the $n$-frame, and the geographic latitude and longitude of the beacon are denoted as $L_t$ and $\lambda_t$, while those of the IMU are denoted as $\hat{L}_b$ and $\hat{\lambda}_b$, respectively.

The Taylor expansion of \eqref{eq:hat_rbe} around the SINS geographic position $\boldsymbol{r}_{b}^n$ is expressed as follows:

\begin{equation}
    \label{eq:rbe_taylor}
    \delta \boldsymbol{r}_b^e 
    = \boldsymbol{H}_b 
    \begin{bmatrix}
        \delta L_b & \delta \lambda_b & \delta h_b \\
    \end{bmatrix}^\top
    = \boldsymbol{H}_b  \delta \boldsymbol{r}_b^n
\end{equation}
where $\delta \boldsymbol{r}_b^n = \left[ \delta L_b, \delta \lambda_b, \delta h_b \right]^\top$ is the SINS geographic position error in the $n$-frame, and $\boldsymbol{H}_b$ is the Jacobian matrix, which can be expressed as follows:

\begin{equation}
    \label{eq:H_b}
    \boldsymbol{H}_b =
    \begin{bmatrix}
        \frac{\partial \boldsymbol{r}_b^e }{ \partial L_b}
        & \frac{\partial \boldsymbol{r}_b^e }{ \partial \lambda_b}
        & \frac{\partial \boldsymbol{r}_b^e }{ \partial h_b} \\
    \end{bmatrix}
\end{equation} 
where the partial derivatives can be calculated as follows:

\begin{equation}
    \label{eq:Hb_partial1}
     \frac{\partial \boldsymbol{r}_b^e }{ \partial L_b} = 
     \begin{bmatrix}
        -\left(R_N + h_b\right) \sin L_b \cos \lambda_b \\
        -\left(R_N + h_b\right) \sin L_b \sin \lambda_b \\
        \left(R_M + h_b\right) \cos L_b
     \end{bmatrix}
\end{equation}

\begin{eqnarray}
    \label{eq:Hb_partial2}
     \frac{\partial \boldsymbol{r}_b^e }{ \partial \lambda_b} = 
     \begin{bmatrix}
        -\left(R_N + h_b\right) \cos L_b \sin \lambda_b \\
        \left(R_N + h_b\right) \cos L_b \cos \lambda_b \\
        0
     \end{bmatrix}
\end{eqnarray}

\begin{equation}
    \label{eq:Hb_partial3}
     \frac{\partial \boldsymbol{r}_b^e }{ \partial h_b} = 
     \begin{bmatrix}
        \cos L_b \cos \lambda_b \\
        \cos L_b \sin \lambda_b \\
        \sin L_b
     \end{bmatrix}
\end{equation}

To ensure consistency with the system's error state definition, the geodetic position errors are transformed into the $n$-frame. \eqref{eq:rbe_taylor} is expressed as follows:

\begin{equation}
    \label{eq:rbe_taylor_n}
    \delta \boldsymbol{r}_b^e 
    = \boldsymbol{H}_b  \delta \boldsymbol{r}_b^n
    = \boldsymbol{H}_b \boldsymbol{D}_r^{-1} \delta \boldsymbol{p}_b^n
\end{equation}
where the transformation matrix can be expressed as 
 $\boldsymbol{D}_r^{-1} = \text{diag} \left(1/\left(R_M + h_b\right), 1/\left[\left(R_N+ h_b\right)\cos L_b \right], -1 \right)$.
 
Substituting \eqref{eq:rbe_taylor_n} into \eqref{eq:hat_pbte} yields:

\begin{equation}
    \label{eq:hat_pbte_final}
    \hat{\boldsymbol{P}}_{bt}^e =
    \boldsymbol{P}_{bt}^e +  \delta \boldsymbol{P}_{bt}^e = 
    \boldsymbol{P}_{bt}^e - \boldsymbol{H}_b \boldsymbol{D}_r^{-1} \delta \boldsymbol{p}_b^n
\end{equation}

Given that the piUSBL installation angles are sufficiently small, the DCM from the $u$-frame to the $b$-frame is linearized as:

\begin{equation}
    \label{eq:hat_cbu}
    \hat{\mathbf{C}}_b^u = \left(\boldsymbol{I} - \delta \boldsymbol{\theta} \times \right) \mathbf{C}_b^u 
\end{equation}

Substituting \eqref{eq:hat_cen}, \eqref{eq:hat_cnb}, \eqref{eq:hat_pbte_final}, and \eqref{eq:hat_cbu} into \eqref{eq_putu_perturbation} yields:

\begin{equation}
    \label{eq:hat_putu_full}
    \begin{aligned}
        \hat{\boldsymbol{P}}_{ut}^u 
        &= \left(\boldsymbol{I} - \delta \boldsymbol{\theta} \times \right) \mathbf{C}_b^u \mathbf{C}_n^b \left(\boldsymbol{I} + \boldsymbol{\phi} \times \right) \left(\mathbf{C}_e^n + \delta \mathbf{C}_e^n\right)  \\
        & \qquad \left(\boldsymbol{P}_{bt}^e +  \delta \boldsymbol{P}_{bt}^e\right) - \left(\boldsymbol{I} - \delta \boldsymbol{\theta} \times \right) \mathbf{C}_b^u \boldsymbol{l}_{bu}^b \\
    \end{aligned}
\end{equation}

Ignoring the second-order error terms, the expression is simplified to:

\begin{equation}
    \label{eq:hat_putu_simplified}
    \begin{aligned}
        \hat{\boldsymbol{P}}_{bt}^u
        & \approx \mathbf{C}_b^u \mathbf{C}_n^b \mathbf{C}_e^n \boldsymbol{P}_{bt}^e - \mathbf{C}_b^u \boldsymbol{l}_{bu}^b + \left( \delta \boldsymbol{\theta} \times \right) \mathbf{C}_b^u \boldsymbol{l}_{bu}^b \\
        & \quad - \left( \delta \boldsymbol{\theta} \times \right) \mathbf{C}_b^u \mathbf{C}_n^b \mathbf{C}_e^n \boldsymbol{P}_{bt}^e + \mathbf{C}_b^u \mathbf{C}_n^b \left(\boldsymbol{\phi} \times \right) \mathbf{C}_e^n \boldsymbol{P}_{bt}^e \\
        & \quad + \mathbf{C}_b^u \mathbf{C}_n^b \delta \mathbf{C}_e^n \boldsymbol{P}_{bt}^e + \mathbf{C}_b^u \mathbf{C}_n^b \mathbf{C}_e^n \delta \boldsymbol{P}_{bt}^e
    \end{aligned}
\end{equation}

By combining the terms and rearranging, we derive:

\begin{equation}
    \label{eq:delta_putu_merged}
    \begin{aligned}
        \delta \boldsymbol{P}_{ut}^u
        & \approx \left[\left(\mathbf{C}_b^u \mathbf{C}_n^b \mathbf{C}_e^n \boldsymbol{P}_{bt}^e\right) \times - \left(\mathbf{C}_b^u \boldsymbol{l}_{bu}^b\right) \times \right] \delta \boldsymbol{\theta} \\
        & \quad - \mathbf{C}_b^u \mathbf{C}_n^b \left[\left(\mathbf{C}_e^n \boldsymbol{P}_{bt}^e\right) \times \right] \boldsymbol{\phi} \\
        & \quad + \mathbf{C}_b^u \mathbf{C}_n^b \left(\delta \mathbf{C}_e^n \boldsymbol{P}_{bt}^e + \mathbf{C}_e^n \delta \boldsymbol{P}_{bt}^e\right) \\
    \end{aligned}
\end{equation}

Based on \eqref{eq:delta_cen} and \eqref{eq:hat_pbte_final}, the error term caused by position uncertainty is further decomposed as:

\begin{equation}
    \label{eq:cen_delta_pbte}
     \mathbf{C}_e^n \delta \boldsymbol{P}_{bt}^e = \mathbf{C}_e^n \left( - \boldsymbol{H}_b \boldsymbol{D}_r^{-1} \delta \boldsymbol{p}_b^n \right) = -  \mathbf{C}_e^n \boldsymbol{H}_b \boldsymbol{D}_r^{-1} \delta \boldsymbol{p}_b^n
\end{equation}

\begin{equation}
    \label{eq:delta_cen_pbte}
    \begin{aligned}
        \delta \mathbf{C}_e^n \boldsymbol{P}_{bt}^e 
        & = \delta \mathbf{C}_e^n 
        \begin{bmatrix}
            x_{bt}^e & y_{bt}^e & z_{bt}^e \\
        \end{bmatrix}^\top \\
        & = \boldsymbol{H}_c
        \begin{bmatrix}
            \delta L_b & \delta \lambda_b & \delta h_b \\
        \end{bmatrix}^\top \\
        & = \boldsymbol{H}_c \boldsymbol{D}_r^{-1} \delta \boldsymbol{p}_b^n
    \end{aligned}
\end{equation}
where $\boldsymbol{H}_c$ can be expressed as \eqref{eq:H_c}.

\begin{figure*}[!h]
\begin{equation}
    \label{eq:H_c}
    \boldsymbol{H}_c = 
    \begin{bmatrix}
        -\cos L_b \cos \lambda_b x_{bt}^e - \cos L_b  \sin \lambda_b y_{bt}^e - \sin L_b z_{bt}^e
        & \sin \lambda_b \sin L_b x_{bt}^e - \cos \lambda_b \sin L_b y_{bt}^e
        & 0 \\
        0 
        & -\cos \lambda_b x_{bt}^e - \sin \lambda_b t_{bt}^e
        & 0 \\
        -\sin L_b \cos \lambda_b x_{bt}^e + \sin L_b  \sin \lambda_b y_{bt}^e - \cos L_b z_{bt}^e
        & \sin \lambda_b \cos L_b x_{bt}^e - \cos \lambda_b \cos L_b y_{bt}^e
        & 0 \\
    \end{bmatrix}
\end{equation}
\end{figure*}

According to the above derivation, \eqref{eq:delta_putu_merged} is reexpressed as:

\begin{equation}
    \label{eq:delta_putu_final}
    \begin{aligned}
        \delta \boldsymbol{P}_{ut}^u
        & \approx \mathbf{C}_b^u \mathbf{C}_n^b \left(\boldsymbol{H}_c - \mathbf{C}_e^n \boldsymbol{H}_b\right) \delta \boldsymbol{p}_b^n \\
        &  \quad - \mathbf{C}_b^u \mathbf{C}_n^b \left[\left(\mathbf{C}_e^n \boldsymbol{P}_{bt}^e\right) \times \right] \boldsymbol{\phi} \\
        &  \quad + \left[\left(\mathbf{C}_b^u \mathbf{C}_n^b \mathbf{C}_e^n \boldsymbol{P}_{bt}^e\right) \times - \left(\mathbf{C}_b^u \boldsymbol{l}_{bu}^b\right) \times \right] \delta \boldsymbol{\theta} \\
    \end{aligned}
\end{equation}

Based on \eqref{eq:measurement_equation}, the measurement innovation of the piUSBL system is further expressed as:

\begin{equation}
    \label{eq:piUSBL_innovation}
    \boldsymbol{Z}_{\text{piUSBL}} = 
    \begin{bmatrix}
        \hat{r}_{\text{SINS}} - \tilde{r}_{\text{piUSBL}} \\
        \hat{\alpha}_{\text{SINS}} - \tilde{\alpha}_{\text{piUSBL}} \\
    \end{bmatrix} = 
    \begin{bmatrix}
        \delta r_{\text{SINS}} - \delta r_{\text{piUSBL}} \\
        \delta \alpha_{\text{SINS}} - \delta \alpha_{\text{piUSBL}} \\
    \end{bmatrix}
\end{equation}
where $\delta r$ and $\delta \alpha$ are the estimated or measured slant range and azimuth errors, respectively.

The piUSBL measurement error can be expressed as:
\begin{equation}
    \label{eq:piUSBL_err}
    \begin{bmatrix}
        \delta r_{\text{piUSBL}} \\
        \delta \alpha_{\text{piUSBL}} \\
    \end{bmatrix} = 
    \begin{bmatrix}
        \delta k r + w_r \\
        w_\alpha \\
    \end{bmatrix}
\end{equation}
where $\delta k$ is the slant range scale factor error, $r$ is the slant range, and $w_r$ and $w_\alpha$ are the measurement noises of the slant range and azimuth, respectively.

According to \eqref{eq:delta_r_alpha} and \eqref{eq:delta_putu_final}, the piUSBL measurement error obtained by the SINS is obtained as follows:

\begin{equation}
    \label{eq:piUSBL_err_sins}
    \begin{bmatrix}
        \delta r_{\text{SINS}} \\
        \delta \alpha_{\text{SINS}} \\
    \end{bmatrix} = 
    \begin{bmatrix}
        \boldsymbol{H}_p & \boldsymbol{H}_\phi & \boldsymbol{H}_\theta\\
    \end{bmatrix}
    \begin{bmatrix}
        \delta \boldsymbol{p}^n \\
        \boldsymbol{\phi} \\
        \delta \boldsymbol{\theta} \\
    \end{bmatrix}
\end{equation}

where $\boldsymbol{H}_p$, $\boldsymbol{H}_\phi$, and $\boldsymbol{H}_\theta$ can be expressed as follows:

\begin{equation}
    \label{eq:Hp_Hphi_Htheta}
    \left\{
    \begin{aligned}
        \boldsymbol{H}_p & = \mathbf{H}_a \mathbf{C}_b^u \mathbf{C}_n^b \left(\boldsymbol{H}_c - \mathbf{C}_e^n \boldsymbol{H}_b\right) \mathbf{D}_r^{-1} \\
        \boldsymbol{H}_\phi & = - \mathbf{H}_a \mathbf{C}_b^u \mathbf{C}_n^b \left[\left(\mathbf{C}_e^n \boldsymbol{P}_{bt}^e\right) \times \right] \\
        \boldsymbol{H}_\theta & = \mathbf{H}_a \left[\left(\mathbf{C}_b^u \mathbf{C}_n^b \mathbf{C}_e^n \boldsymbol{P}_{bt}^e\right) \times - \left(\mathbf{C}_b^u \boldsymbol{l}_{bu}^b\right) \times \right]
    \end{aligned}
    \right.
\end{equation}

Substituting \eqref{eq:piUSBL_err} and \eqref{eq:piUSBL_err_sins} into \eqref{eq:piUSBL_innovation}, we can obtain the piUSBL measurement innovation. Besides, the innovation of the depth gauge can be expressed as:
\begin{equation}
    \label{eq:depth_innovation}
    \begin{aligned}
        Z_{\text{Depth}} 
        & = \hat{h}_{\text{SINS}}^n - \tilde{h}_{\text{Depth}}^n \\
        & = h_{\text{SINS}}^n + \boldsymbol{e}_3 \left( \delta \boldsymbol{p}^n + \left[\boldsymbol{I} - \left(\boldsymbol{\phi} \times \right)\right] \mathbf{C}_b^n \boldsymbol{l}_{bd}^b \right)\\
        & \quad - \left(h_{\text{Depth}}^n + \delta h^n + w_h \right) \\
        & =\boldsymbol{e}_3 \delta \boldsymbol{p}^n + \boldsymbol{e}_3 \left[\left(\mathbf{C}_b^n \boldsymbol{l}_{bd}^b\right) \times \right] \boldsymbol{\phi} - \delta h^n - w_h \\
    \end{aligned}
\end{equation}
where $\boldsymbol{e}_3 = \left[0,0,1\right]$, and the expression can be further rearranged to the depth error as follows:

\begin{equation}
    \label{eq:depth_err}
    \delta h_{\text{SINS}} = 
    \begin{bmatrix}
        \boldsymbol{H}_{h} & \boldsymbol{H}_{l} \\
    \end{bmatrix}
    \begin{bmatrix}
        \delta \boldsymbol{p}^n \\
        \boldsymbol{\phi} \\
    \end{bmatrix}
\end{equation}
where $\boldsymbol{H}_{h}$ and $\boldsymbol{H}_{l}$ can be expressed as:
\begin{equation}
    \label{eq:Hh_Hl}
    \left\{
    \begin{aligned}
        \boldsymbol{H}_{h} & = \boldsymbol{e}_3 \\
        \boldsymbol{H}_{l} & = \boldsymbol{e}_3 \left[\left(\mathbf{C}_b^n \boldsymbol{l}_{bd}^b\right) \times \right]
    \end{aligned}
    \right.
\end{equation} 

In summary, the measurement matrix $\boldsymbol{H}$ in the tightly coupled navigation system is expressed as:

\begin{equation}
    \label{eq:measurement_transition_matrix}
    \begin{aligned}
        \boldsymbol{H} 
        &= 
        \begin{bmatrix}
            \boldsymbol{H}_{\text{piUSBL}} \\
            \boldsymbol{H}_{\text{Depth}} \\
        \end{bmatrix}_{(3 \times 26)} \\
        & =
        \begin{bmatrix}
            \boldsymbol{H}_p & \boldsymbol{0}_{2\times 3} & \boldsymbol{H}_\phi & \boldsymbol{0}_{2\times 12} & \boldsymbol{H}_\theta & -\boldsymbol{r} & 0\\
            \boldsymbol{H}_h & \boldsymbol{0}_{1\times 3} & \boldsymbol{H}_l & \boldsymbol{0}_{1\times 12} & \boldsymbol{0}_{1\times 3} & 0 & -1 \\
        \end{bmatrix}
    \end{aligned}
\end{equation}
where $\boldsymbol{r} = [r, 0]^\top$ is formed to match the dimension of the measurement vector, and $r$ denotes the piUSBL slant-range measurement.

The above is the derivation process of the tightly coupled navigation system. Compared with the loosely coupled navigation system, the tightly coupled navigation system converts the observation function from the position measurement to basic geometric measurements, like slant range and azimuth, which can avoid solution failure resulting from the loss or degradation of any individual measurement.

\subsection{Observability Analysis}
The local weak observability of the proposed tightly coupled navigation system under nominal sensor availability is analyzed in this subsection.
Since the error state $\boldsymbol{X}$ in \eqref{eq:error_state} includes both navigation errors and slowly varying sensor/calibration errors, the analysis focuses on the navigation-relevant states, i.e., $\delta \boldsymbol{p}^n$, $\delta \boldsymbol{v}^n$, and $\boldsymbol{\phi}$, which directly determine the navigation performance.

Along the estimated trajectory, the linearized error dynamics can be written in discrete form as
\begin{equation}
\label{eq:linear_discrete_err}
\boldsymbol{X}_{k+1} = \boldsymbol{\Phi}_k \boldsymbol{X}_k, \qquad
\boldsymbol{Z}_k = \boldsymbol{H}_k \boldsymbol{X}_k + \boldsymbol{V}_k
\end{equation}
where $\boldsymbol{\Phi}_k$ is obtained by discretizing the state transition matrix $\boldsymbol{F}$ in \eqref{eq:state_transition_matrix}, and $\boldsymbol{H}_k$ is given by the measurement transition matrix derived in \eqref{eq:measurement_transition_matrix}.

Over a finite horizon of $N$ measurement epochs, the time-varying observability matrix $\mathcal{O}_{k,N}$ is constructed as
\begin{equation}
\label{eq:observability_matrix}
\mathcal{O}_{k,N} =
\begin{bmatrix}
\boldsymbol{H}_k\\
\boldsymbol{H}_{k-1}\boldsymbol{\Phi}_{k-1}\\
\boldsymbol{H}_{k-2}\boldsymbol{\Phi}_{k-2}\boldsymbol{\Phi}_{k-1}\\
\vdots\\
\boldsymbol{H}_{k-N+1}\boldsymbol{\Phi}_{k-N+1}\cdots\boldsymbol{\Phi}_{k-1}
\end{bmatrix}
\end{equation}
where $k$ denotes the measurement-epoch index and $N$ is the finite-horizon window length (in measurement epochs).
The local weak observability is evaluated using the numerical rank of $\mathcal{O}_{k,N}$ and its singular values.

To explicitly assess the observability of the navigation-relevant states, the submatrix $\mathcal{O}^{\text{nav}}_{k,N}$ is formed by selecting the columns of $\mathcal{O}_{k,N}$ corresponding to $\{\delta \boldsymbol{p}^n,\delta \boldsymbol{v}^n,\boldsymbol{\phi}\}$ in \eqref{eq:error_state}.
Under nominal AUV motion and nominal availability of the minimal observations $(r,\alpha,h^n)$ in \eqref{eq:measurement_equation}, $\mathcal{O}^{\text{nav}}_{k,N}$ does not exhibit numerical rank deficiency over the considered horizon, indicating that the position, velocity, and attitude errors are locally weakly observable.

Although the piUSBL provides only the slant range and azimuth measurements, the depth-gauge innovation in \eqref{eq:depth_innovation} introduces nonzero sensitivity to $\delta \boldsymbol{p}^n$ and $\boldsymbol{\phi}$ through the corresponding Jacobian terms, thereby providing complementary constraints that mitigate the limited vertical-angle information of the planar piUSBL configuration.
This analysis supports the claim that the proposed compact observation set \eqref{eq:measurement_equation} is sufficient to maintain local weak observability of the navigation-relevant states under nominal motion.

\section{Simulation and Field Experiments}
\label{sec:simulation_and_field_experiments}

To analyze the impact of time delay in underwater acoustic positioning sensors, and to validate both the proposed time-delay measurement method and the effectiveness of the proposed system, simulations and field experiments were conducted. 

\subsection{Simulation}

\begin{figure}[!t]
    \centering
    \includegraphics[width=\linewidth]{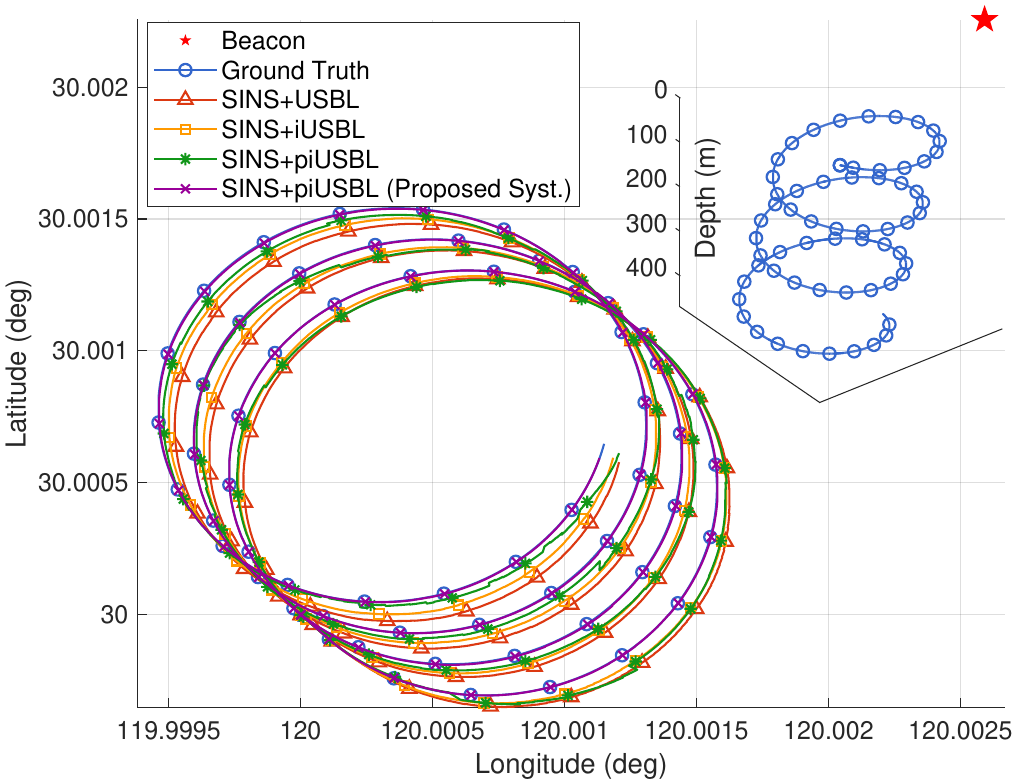}
    \caption{Simulation trajectory curves.}
    \label{fig:simulation_trace}
\end{figure}

To investigate the impact of time delay on performance in USBL, iUSBL, and piUSBL systems and the proposed tightly coupled navigation method, a series of simulation experiments were designed, as illustrated in Fig. \ref{fig:simulation_trace}.

The simulation trajectory follows a helical descent path, starting from the position $\left(30^\circ, 120^\circ, 0 \ \text{m}\right)$. The AUV descends along a pitch angle of $-15^\circ$, accelerating to 6 knots, and finally descending to a depth of 470 m. The beacon is positioned 250 m east and 250 m north of the origin. This simulation scenario effectively replicates a realistic underwater operation profile for an AUV during a mission involving continuous descent and motion. The sensor specifications used in the simulation are summarized in Table \ref{tab_simSpec}.

\begin{table}
  \begin{center}
    \caption{Simulation Specifications}
    \label{tab_simSpec}
    \begin{tabular}{l c r}
      \toprule
      Equipment & Specification & Accuracy \\
      \midrule
      Gyroscope & Constant bias & 0.01 deg/h \\
        & Random walk & 0.01 deg/$\sqrt{\mathrm{h}}$ \\
        & Scale factor error & 50 ppm \\
      Accelerometer & Constant bias & 50  $\mu$g \\
        & Random walk & 0.01 m/s/$\sqrt{\mathrm{h}}$ \\
        & Scale factor error & 100 ppm \\
      USBL/iUSBL & Axis-$x$ angle error & 0.1 $\deg$ \\
        & Axis-$y$ angle error & 0.1 $\deg$ \\
        & Slant range (SR) error & 0.1 \% SR  \\
        &  Measurement frequency & 1 Hz \\
      piUSBL & Azimuth error & 0.1 $\deg$ \\
        & Slant range (SR) error & 0.1 \% SR  \\
        &  Measurement frequency & 1 Hz \\
      \bottomrule
    \end{tabular}
  \end{center}
\end{table}

\begin{figure}[!t]
    \centering
    \includegraphics[width=\linewidth]{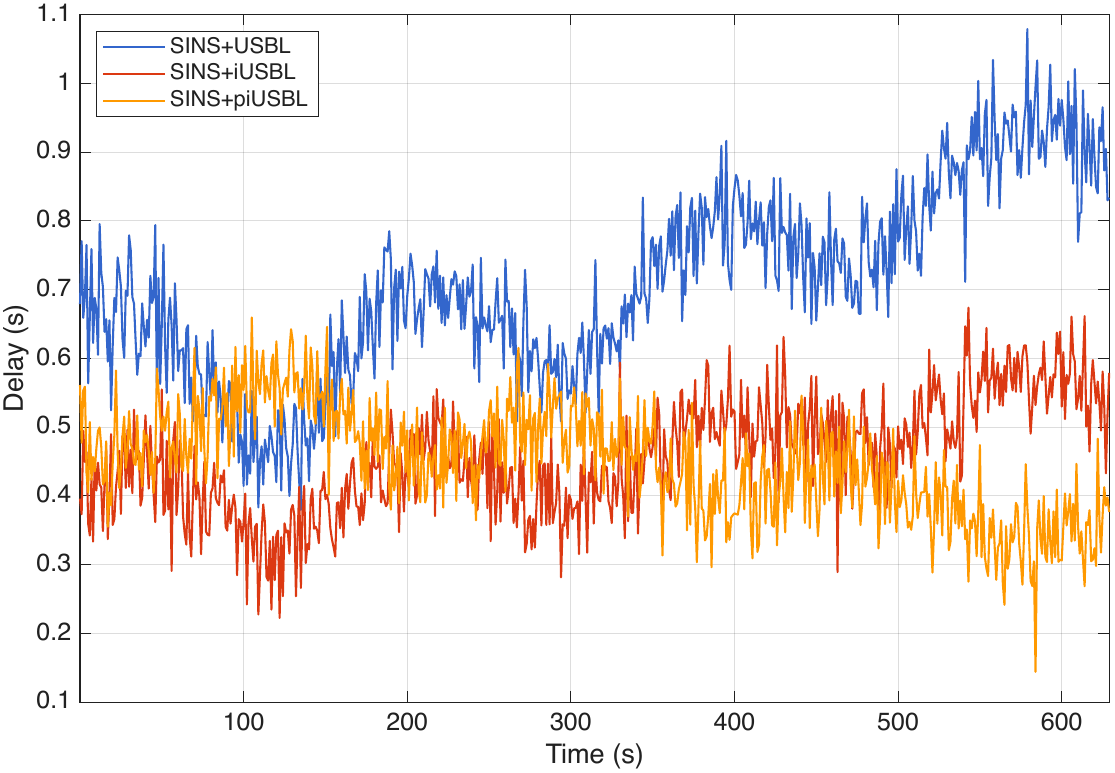}
    \caption{Time delay error of the USBL, iUSBL, and piUSBL systems in simulation.}
    \label{fig:simulationDelayError}
\end{figure}

It is worth noting that, in this simulation, we considered the acoustic communication delay, the positioning signal delay, and the processing delay of the USBL system, which corresponding to the $t_1-t_4$ in Fig. \ref{fig:SystemInfo}. Due to the inverted structure of the iUSBL and piUSBL systems, the acoustic communication delay is not considered in the iUSBL and piUSBL systems, as illustrated in Fig. \ref{fig:simulationDelayError}.
The USBL system exhibits a time delay that increases significantly with beacon distance, while the iUSBL system shows lower delays, though with a similarly increasing trend.
In comparison, the piUSBL system has a similar delay magnitude to iUSBL, but due to the use of a sampling window (see Eq. \eqref{eq:delay_definition}), the closer the signal arrival is to the end of the window, the smaller the delay—leading to an inverse trend relative to the iUSBL system.
The USBL and piUSBL systems are tightly coupled with the SINS. The structure of the system can be found in \cite{wangRobustFilterMethod2023}.

\begin{table}
    \centering
    \caption{Position error of the four systems in simulation.}
    \label{tab_simulationPosError}
    \begin{tabular}{lcccccc}
    \toprule
    \multirow{2}{*}{\textbf{Sub Syst.}} & \multicolumn{3}{c}{RMSE(m)} & \multicolumn{3}{c}{MAXERR(m)} \\
    \cline{2-7}
    & $e_N$ & $e_E$ & $e_D$ & $e_N$ & $e_E$ & $e_D$ \\
    \midrule
    USBL & 5.65 & 5.06 & 0.55 & 8.41 & 6.59 & 0.76 \\
    iUSBL & 3.78 & 3.42 & 0.35 & 5.55 & 4.41 & 0.47 \\
    piUSBL & 3.12 & 3.14 & 0.42 & 6.37 & 6.65 & 3.23 \\
    Prop. Syst. & 0.17 & 0.28 & 0.02 & 0.42 & 0.48 & 0.16 \\
    \bottomrule
\end{tabular}
\end{table}

\begin{figure}[!t]
    \centering
    \includegraphics[width=\linewidth]{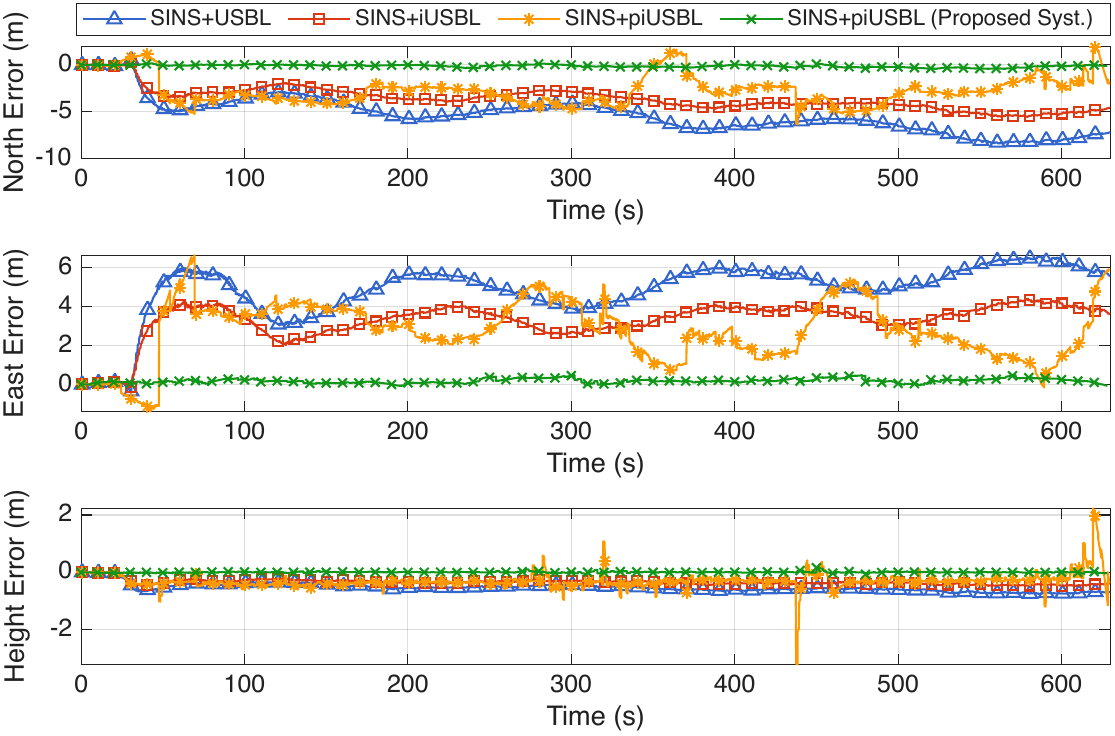}
    \caption{Position error of the SINS/USBL, SINS/USBL, SINS/piUSBL, and proposed SINS/piUSBL with time delay accurate measurement aided integrated navigation system in simulation.}
    \label{fig:simulationPosError}
\end{figure}

The simulation results, presented in Fig. \ref{fig:simulationPosError} and Table \ref{tab_simulationPosError}, show that the SINS/USBL system exhibits the largest position error, both in terms of root-mean-square error (RMSE) and maximum error (MAXERR), across the North, East, and Height/Down projections. This is primarily due to the significant time delay associated with the USBL system, compared to the iUSBL and piUSBL systems.

The SINS/iUSBL system demonstrates a smaller position error by eliminating the need for acoustic communication delay. Compared to the SINS/USBL system, the RMSE of the SINS/iUSBL system is reduced by 32.89\%, from 7.61 m to 5.11 m, and the MAXERR is reduced by 33.65\%, from 10.91 m to 7.11 m. It is important to note that, unless otherwise specified, both the RMSE and MAXERR are computed using the Euclidean norm.

The SINS/piUSBL system shows a slight reduction in position error in terms of RMSE, with a 41.52\% reduction from 7.61 m to 4.45 m compared with SINS/USBL system. However, it is important to note that the SINS/piUSBL system has a larger MAXERR than the SINS/iUSBL system. This is because the piUSBL system lacks an angular measurement, which reduces its robustness to external disturbances and reveals a higher susceptible to time delay. As a result, when errors occur in the observations, the system's filter is destabilized, leading to fluctuations in the estimated position.

The proposed SINS/piUSBL system with accurate time delay measurement, referred to as the Prop. Syst. in Table \ref{tab_simulationPosError}, demonstrates the smallest position error. The RMSE is reduced by 96.45\%, from 7.61 m to 0.27 m, and the MAXERR is reduced by 93.95\%, from 10.91 m to 0.66 m, compared with the SINS/USBL system.

In summary, the piUSBL-based integrated navigation system outperforms both the USBL and iUSBL systems in terms of average error, despite exhibiting significant oscillations when handling outliers caused by time delay. By introducing the proposed time delay error estimation method, this issue is effectively addressed, resulting in a significant improvement in the accuracy of the compensated piUSBL system.

\subsection{Field Experiments}

To further validate the performance of the proposed system and assess the effectiveness of accurate time delay measurement, field experiments were conducted. 

\begin{figure}[!t]
    \centering
    \includegraphics[width=\linewidth]{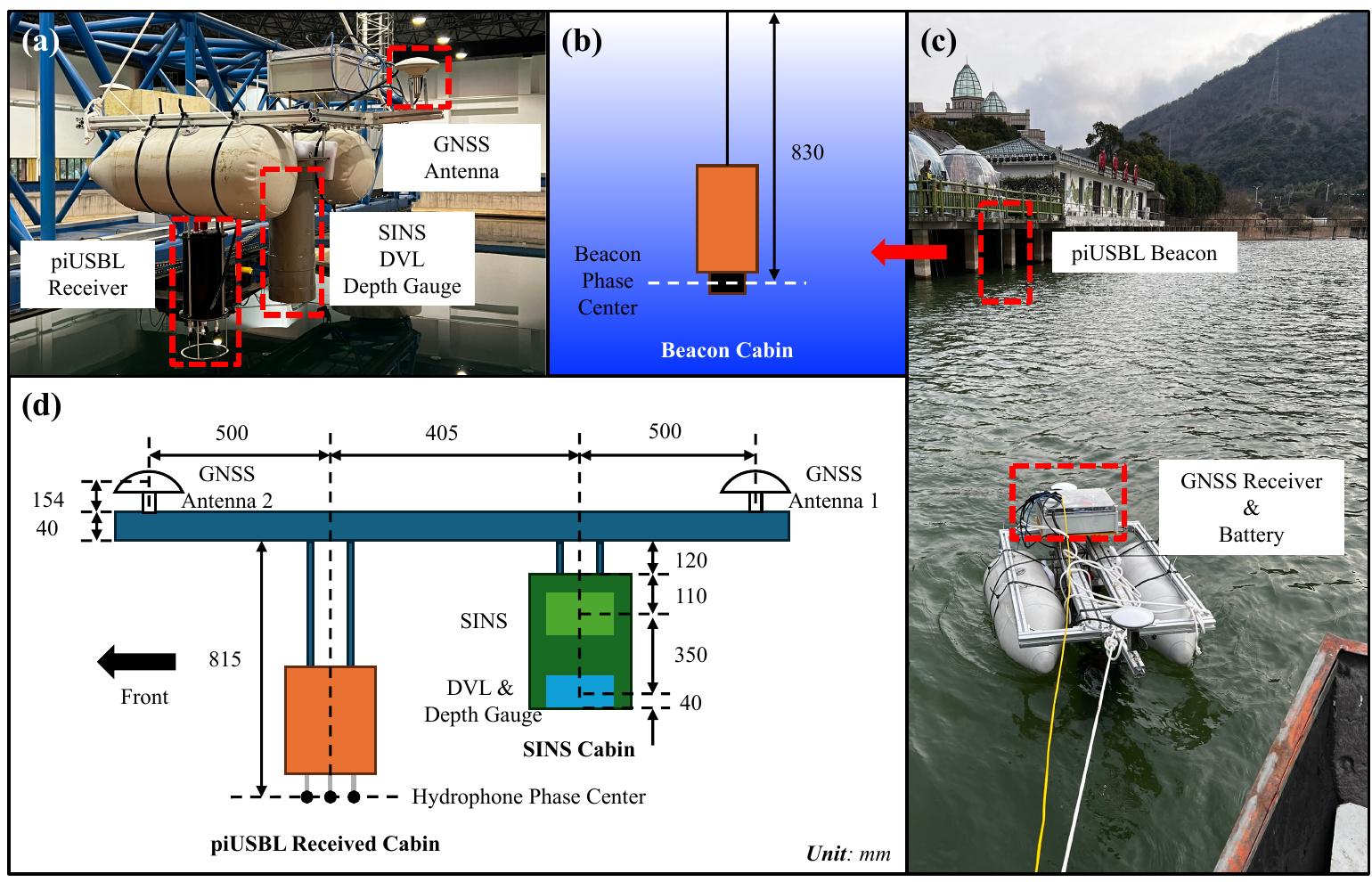}
    \caption{Setup of the SINS/piUSBL system during field experiments.}
    \label{fig:field_exp_structure}
\end{figure}

The field experiment was conducted at the Tannong Reservoir, and its system structure is shown in Fig. \ref{fig:field_exp_structure}.
In the experimental setup, the acoustic beacon was fixed on the shore at a depth of 0.83 m underwater. The experimental platform was equipped with the piUSBL receiver system, SINS, DVL, depth gauge, and Differential Global Navigation Satellite System (DGNSS). The sensor performance parameters are provided in Table \ref{tab_expSpec}. 

The relative positions of the systems, as shown in Fig. \ref{fig:field_exp_structure}(d), were accurately measured, with the values expressed in millimeters. 
The experiment was conducted by towing the platform with a small boat, as illustrated in Fig. \ref{fig:field_exp_structure}(c).

\begin{table}
  \begin{center}
    \caption{Field Experiment Specifications}
    \label{tab_expSpec}
    \begin{tabular}{l c r}
      \toprule
      Equipment & Specification & Accuracy \\
      \midrule
      Gyroscope & Constant bias & 0.03 deg/h \\
        & Random walk & 0.005 deg/$\sqrt{\mathrm{h}}$ \\
        & Scale factor error & 100 ppm \\
      Accelerometer & Constant bias & 300 $\mu$g \\
        & Random walk & 0.05 m/s/$\sqrt{\mathrm{h}}$ \\
        & Scale factor error & 200 ppm \\
      piUSBL & Depth error & 0.1 $\deg$ \\
        &  Slant range (SR) error & 0.1\% SR \\
        &  Measurement frequency & 1 Hz \\
      OCXO & Clock drift ($\le$ 2 hr) & 3 $\mathrm{\mu} $s/hr \\
        & Clock drift ($>$ 2 hr) & 18 $\mu$s/hr \\
      Depth gauge & Depth error & 0.1 m \\
      DGNSS & Position error (CORS) & 0.02 m \\
      \bottomrule
    \end{tabular}
  \end{center}
\end{table}

Before the experiment, the OCXOs onboard the piUSBL system’s transmitter and receiver were disciplined using a high-precision GNSS clock for 30 minutes. 
Based on the clock performance we employed, the time offset between the two clocks remained below 6 $\mu$s throughout the experiment.
When the typical sound speed in freshwater is around 1500 m/s, the slant range error caused by the clock offset is less than 0.009 m, which is negligible compared to the piUSBL system's inherent measurement error.

\begin{figure}[!t]
    \centering
    \includegraphics[width=\linewidth]{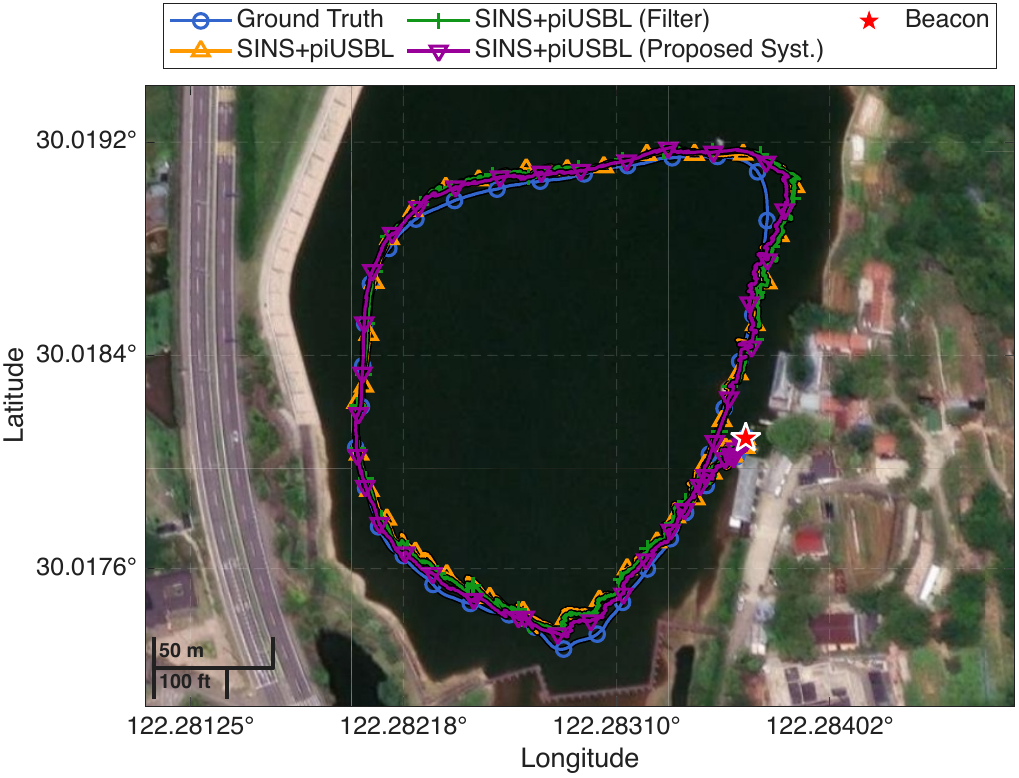}
    \caption{Trajectory of the ground truth, SINS/piUSBL, SINS/piUSBL with filter-based delay compensation, and SINS/piUSBL with time delay accurate measurement aided integrated navigation system in field experiment.}
    \label{fig:field_exp_trace}
\end{figure}

The experimental trajectory is shown in Fig. \ref{fig:field_exp_trace}. 
To evaluate navigation performance, we implemented both the conventional SINS/piUSBL tightly coupled method, the filter-based time delay compensation method, and the proposed tightly coupled navigation system based on precise time-delay measurements.
The filter-based time delay compensation method treats the time delay as an additional state in the filter, estimating it alongside other system states.
This approach allows the filter to adaptively compensate for time delay variations during operation, which is a commonly used technique in delay compensation for integrated navigation systems \cite{leeStateParameterEstimation2017}.

\begin{figure}[!t]
    \centering
    \includegraphics[width=\linewidth]{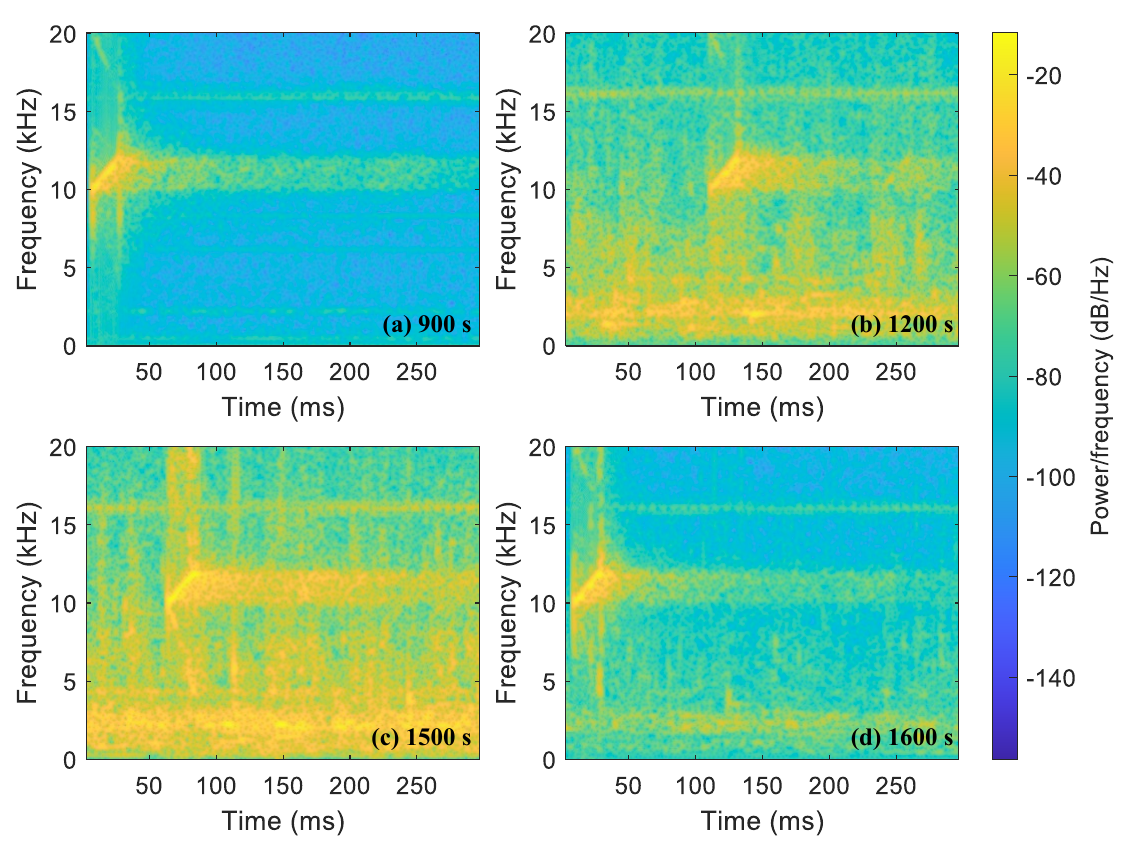}
    \caption{Spectrogram of the piUSBL signal received in field experiments.}
    \label{fig:field_exp_spectrogram}
\end{figure}

Based on site observations, the north and south sides of the reservoir are shallower and exhibit more complex acoustic conditions, which can degrade the piUSBL measurements.
Fig. \ref{fig:field_exp_spectrogram} presents the spectrogram of the piUSBL signal recorded during the field experiments. 
Subplots (a)–(d) correspond to the time intervals centered at 900, 1200, 1500, and 1600 s, respectively. 
Subplots (a) and (d) exhibit clearer signal features, suggesting more favorable acoustic conditions. 
In contrast, subplots (b) and (c) show pronounced interference from environmental noise, multipath, and reverberation, indicating degraded acoustic conditions, consistent with our observations at the experimental site.
To further quantify the measurement quality, we computed the SNR of the piUSBL signal over the analyzed segment. 
The SNR ranged from 8.70 dB to 59.15 dB, with a mean of 29.56 dB, indicating substantial variability in acoustic measurement conditions during the experiment.
This variability is consistent with the spectrogram observations in Fig. \ref{fig:field_exp_spectrogram}, where certain intervals exhibit stronger noise, multipath, and reverberation interference.
In degraded acoustic environments, the reduced reliability of TOF estimation and time-delay measurement induces a larger uncertainty in the effective measurement epoch $t_1$. 
The system is designed to encode this uncertainty by inflating the measurement covariance $\boldsymbol{R}$, which appropriately down-weights the acoustic update and prevents the disturbance on the trajectory.

\begin{figure}[!t]
    \centering
    \includegraphics[width=\linewidth]{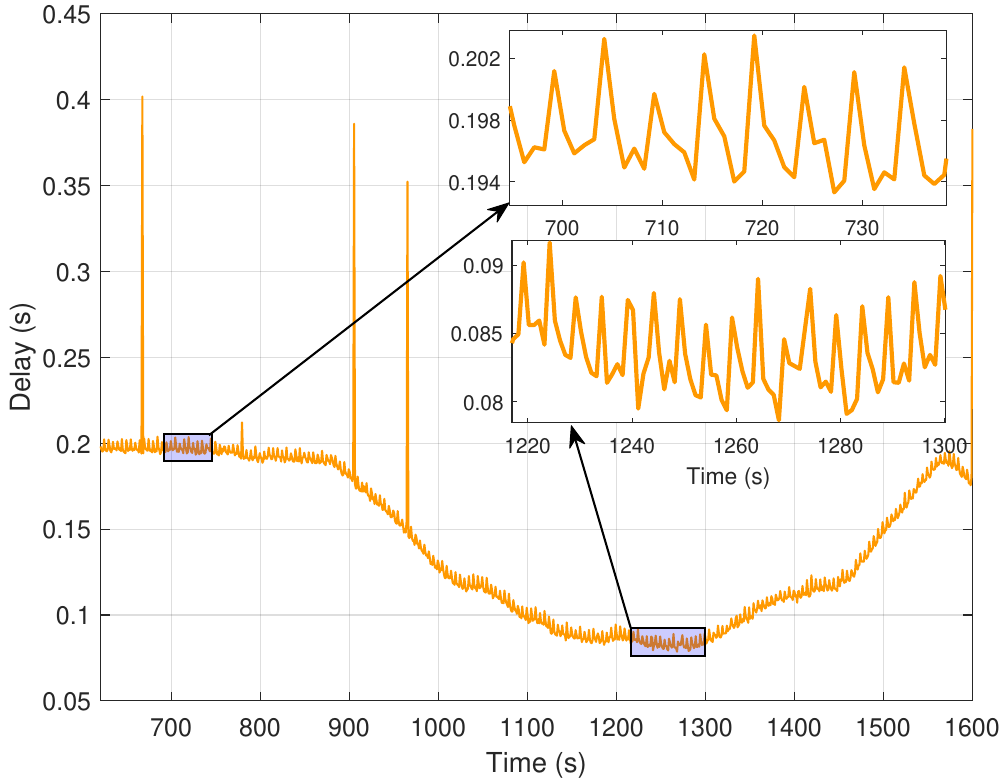}
    \caption{Time delay error of the piUSBL systems measured in field experiments.}
    \label{fig:field_exp_delay_err}
\end{figure}

Fig. \ref{fig:field_exp_delay_err} shows the time delay measured using the synchronized clocks. The results clearly indicate that the delay is influenced by system processing delays and other factors, maintaining an amplitude of approximately 8 ms across different parts of the system. Additionally, the delay is observed to vary with the distance from the acoustic source. Furthermore, due to issues such as instability in serial communication, noticeable jumps in the delay are also evident.
Based on these delay measurements, it is apparent that when a commonly used 200 Hz IMU is employed as the sensor in the INS, the presence of the delay results in piUSBL observations being delayed by tens of inertial measurement updates, which significantly impacts the accuracy of the system.

\begin{figure}[!t]
    \centering
    \includegraphics[width=\linewidth]{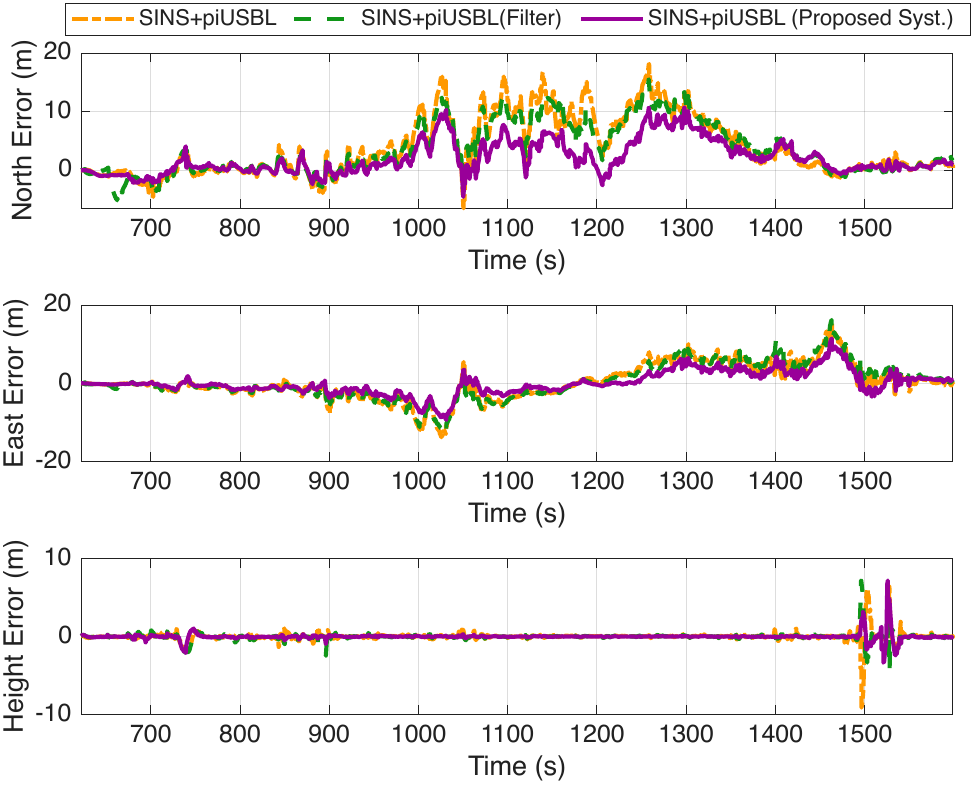}
    \caption{Position error of the SINS/piUSBL, SINS/piUSBL with filter-based delay compensation, and SINS/piUSBL with time delay accurate measurement aided integrated navigation system in field experiment.}
    \label{fig:field_exp_pos_err}
\end{figure}

Fig. \ref{fig:field_exp_pos_err} shows the position error trajectories in the navigation frame for three integration approaches: conventional SINS/piUSBL tightly coupled, SINS/piUSBL with filter-based delay compensation, and the proposed SINS/piUSBL with accurate time delay measurement-aided compensation.
The ground truth is obtained using the DGNSS system, which uses the continuous operating reference stations (CORS) to provide high-precision position information.

The results reveal that the filter-based delay compensation provides improvement over the baseline, while the proposed accurate time delay measurement method achieves substantially higher accuracy by effectively reducing the position error throughout the trajectory.

It is worth noting that the error peaks in Fig. \ref{fig:field_exp_pos_err} do not necessarily coincide with the minimum residual system delay in Fig. \ref{fig:field_exp_delay_err}. 
A small residual delay indicates improved time alignment; however, the instantaneous positioning accuracy also depends on acoustic measurement quality and vehicle maneuvers. In our field trial, the increased north error around 1000–1300 s is primarily attributable to degraded acoustic conditions, as evidenced by Fig. \ref{fig:field_exp_spectrogram}, which weakens the acoustic position constraints. 
The pronounced height fluctuation near 1500 s is consistent with a transient vertical motion of the vehicle, which excites the vertical channel and leads to a short-term error increase. 
Consequently, although time-delay compensation mitigates a major error source, segment-wise performance can still be dominated by other practical disturbances in field conditions.

\begin{figure}[!t]
    \centering
    \includegraphics[width=\linewidth]{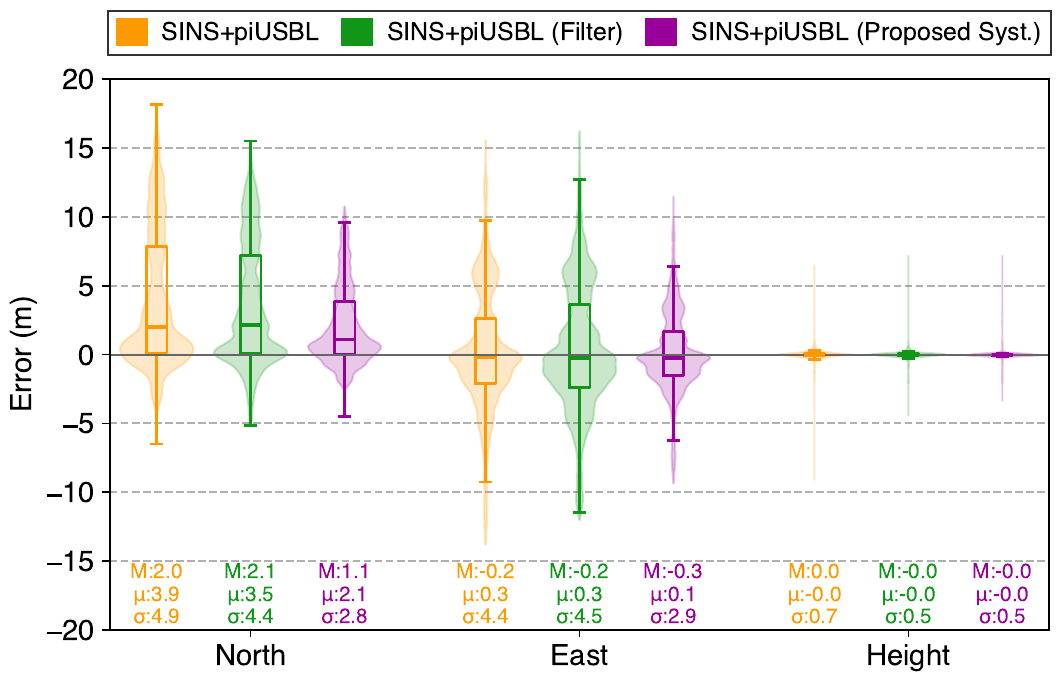}
    \caption{Violin plot of position error distribution in the north, east, and height directions comparing SINS/piUSBL, SINS/piUSBL with filter-based delay compensation, and SINS/piUSBL with time delay accurate measurement aided compensation in field experiment ($M$, $\mu$, and $\sigma$ denote median, mean, and standard deviation, respectively).}
    \label{fig:field_exp_pos_err_violin}
\end{figure}

\begin{table*}[!h]
    \centering
    \caption{Position error comparison: SINS/piUSBL without compensation, with filter-based compensation, and with proposed accurate time delay measurement aided compensation in field experiment.}
    \label{tab_positionError}
    \begin{tabular}{lcccccc}
    \toprule
    \multirow{2}{*}{\textbf{Method}} & \multicolumn{3}{c}{RMSE (m)} & \multicolumn{3}{c}{MAXERR (m)} \\
    \cline{2-7}
    & $e_N$ & $e_E$ & $e_D$ & $e_N$ & $e_E$ & $e_D$ \\
    \midrule
    SINS/piUSBL tightly coupled (no comp.) & 6.27 & 4.44 & 0.68 & 18.14 & 15.57 & 9.11 \\
    Filter-based compensation & 5.63 & 4.55 & 0.51 & 15.51 & 16.22 & 7.15 \\
    \textbf{Proposed system (Delay Meas.)}  & \textbf{3.51} & \textbf{2.92} & \textbf{0.45} & \textbf{10.74} & \textbf{11.47} & \textbf{7.15} \\
    \cmidrule(lr){1-7}
    Error reduction vs. baseline (\%) & 44.02 & 34.23 & 33.82 & 40.79 & 26.33 & 21.51 \\
    Error reduction vs. filter-based (\%) & 37.66 & 35.82 & 11.76 & 30.69 & 29.31 & 0.00 \\
    \bottomrule
    \end{tabular}
\end{table*}
 
Fig. \ref{fig:field_exp_pos_err_violin} compares the position error distribution of the tightly coupled SINS/piUSBL baseline, the filter-based delay compensation method, and the proposed delay-measurement-aided approach.
As shown by the violin plots, the proposed approach yields noticeably more concentrated error distributions in both horizontal components, with reduced dispersion and fewer large-error tails, indicating improved robustness under delay-varying conditions.

The quantitative results are summarized in Table \ref{tab_positionError}.
Compared with the baseline without compensation, the proposed approach reduces the RMSE by 6.27, 4.44, and 0.68 m to 3.51, 2.92, and 0.45 m in the north, east, and down directions, respectively, which corresponds to RMSE reductions of 44.02\%, 34.23\%, and 33.82\%.
The maximum error decreases from 18.14, 15.57, 9.11 m to 10.74, 11.47, 7.15 m, corresponding to reductions of 40.79\%, 26.33\%, and 21.51\%.
Relative to the filter-based compensation, the proposed approach further improves the horizontal RMSE by 37.66\% (North) and 35.82\% (East), and reduces the horizontal maximum error by 30.69\% and 29.31\%, respectively, while achieving comparable vertical maximum error.

Field experiments validate the proposed SINS/piUSBL tightly coupled navigation system under minimal sensing constraints, including cases where vertical-angle observation is unavailable with a planar-array configuration.
The OCXO-based time delay measurement method provides reliable end-to-end delay estimates and enables online compensation within the fusion architecture.
As shown in Fig. \ref{fig:field_exp_pos_err_violin} and Table \ref{tab_positionError}, the proposed approach substantially improves positioning accuracy compared with the uncompensated baseline, reducing both RMSE and maximum error in the horizontal and vertical components. 
In the field experiments, it also outperforms the filter-based delay compensation in the horizontal directions, while achieving comparable performance in the vertical direction. 
The improvement trend relative to the uncompensated baseline is consistent with the simulation results.

\section{Conclusions}
\label{sec:conclusion}

Time delays in multi-sensor fusion can introduce non-negligible errors in underwater navigation, especially when the effective measurement epoch is misaligned with the fusion epoch. 
To address this issue, this paper proposed a tightly coupled SINS/piUSBL/depth-gauge integrated navigation system with precise time synchronization and delay-aware processing. 
Rather than treating the delay as a nuisance parameter to be estimated, the proposed architecture exploits synchronized timing in the piUSBL processing chain to explicitly quantify the end-to-end delay and to perform the measurement update at the corresponding effective measurement epoch, followed by state propagation to the current output time. 
The proposed framework was evaluated through simulations and validated in field experiments. 
The main conclusions are summarized as follows:

\begin{itemize}
    \item [1.] \textbf{Impact of delay-induced epoch mismatch}: This paper systematically analyzed how time delays affect acoustic-aided navigation accuracy. The results show that delay-induced errors mainly originate from the mismatch between the effective measurement epoch and the fusion epoch. Comparative simulations of USBL, iUSBL, and piUSBL further indicate that iUSBL and piUSBL exhibit smaller intrinsic delays than conventional USBL under their respective operating mechanisms, with piUSBL presenting the lowest intrinsic delay in our setup. The simulations also highlight that, while piUSBL can provide competitive average positioning accuracy, its performance can be affected by the lack of vertical-angle information inherent to planar-array configurations.
    \item [2.] \textbf{Tightly coupled minimal-observation fusion with observability considerations}: Based on the self-developed piUSBL system, this paper formulated a compact tightly coupled fusion scheme that integrates piUSBL azimuth and slant range with depth-gauge measurements as complementary observations. This design reduces the dependence on an additional angular measurement compared with traditional acoustic configurations while maintaining local weak observability under nominal sensor availability, providing a practical reference for deploying low-complexity acoustic sensing in integrated navigation.
    \item [3.] \textbf{Delay measurement and compensation enabled by synchronized timing}: By integrating synchronized timing into both the acoustic processing chain and the fusion architecture, the proposed method measures the end-to-end delay and compensates for both acoustic propagation and system processing delays in an online manner. Field experiments demonstrate that the proposed delay-aware fusion reduces the position RMSE by 44.02\% and decreases the MAXERR by 32.55\% compared with the case without delay compensation, consistent with the trends observed in simulation. Furthermore, when compared with the filter-based delay compensation method, the proposed approach achieves additional RMSE reductions of 37.66\% and 35.82\% in the horizontal directions, demonstrating the advantage of explicit delay measurement over filter-based delay estimation. The field trials also reveal that while delay compensation substantially improves overall accuracy, instantaneous positioning performance remains sensitive to acoustic measurement quality, as evidenced by increased errors during periods of degraded SNR and multipath interference, highlighting the importance of robust acoustic signal processing in conjunction with accurate time synchronization.
\end{itemize}

Due to limitations of the current experimental conditions and equipment, further validation under more diverse operational scenarios is desirable. 
Future work will extend the proposed delay-aware fusion architecture to a broader range of acoustic-aided navigation configurations and investigate deeper integration of acoustic raw-data features with inertial navigation, with the goal of improving robustness under challenging channel conditions and vehicle dynamics.

{\appendix[Pseudocode of the Tightly Coupled SINS/piUSBL/Depth-Gauge System]

In implementation, the estimator operates in an event-driven manner with asynchronous sensor updates.
The IMU provides high-rate inertial increments to propagate the navigation state, while the depth gauge provides high-rate measurements that are processed independently to continuously constrain the vertical channel.
Each piUSBL update is processed jointly with a depth observation synchronized to the effective piUSBL measurement epoch.

Since the depth-gauge rate is typically much higher than that of the piUSBL, the depth at a desired epoch is obtained from the depth buffer by interpolation using adjacent depth samples.
When time-delay compensation is enabled, the piUSBL measurement is associated with an explicitly measured end-to-end delay $\Delta t$, and its effective measurement time is mapped to $t_1=t_4-\Delta t$, where $t_4$ denotes the current processing time (the latest IMU epoch).
The buffered SINS state $\boldsymbol{X}(t_1)$ is retrieved for the joint update, and the state is then propagated from $t_1$ to $t_4$ using the buffered IMU data, which is equivalent to a filter re-calculation (FR) over the interval $[t_1,t_4]$.

To robustly handle abrupt changes of the effective measurement epoch, an epoch-mismatch threshold tied to the IMU update interval is applied.
Let $\Delta t_{\mathrm{IMU}}$ be the nominal IMU update interval and $e=|t_4-t_1|$ be the epoch mismatch for the current piUSBL update.
If $e<0.1\,\Delta t_{\mathrm{IMU}}$, the piUSBL update is treated as synchronous with the current epoch and applied at $t_4$ for computational simplicity.
Otherwise, the compensation at the current epoch is suspended and a dedicated FR thread is triggered: the joint update is performed at $t_1$ using buffered data, and the updated estimate is re-propagated to $t_4$ using the buffered IMU measurements.
The FR result at $t_4$ then replaces the current estimate, ensuring consistent timing in the tightly coupled filter.
Note that historical information is used deterministically through the FR mechanism (buffered IMU replay) rather than to predict the delay, because $\Delta t$ is explicitly measured by the synchronized timing support.
The main processing loop is summarized in Algorithm~\ref{alg_mainloop}.

\begin{algorithm}[!h]
\caption{Main Processing Loop of the Tightly Coupled SINS/piUSBL/Depth-Gauge System}
\label{alg_mainloop}
\begin{algorithmic}[1]
\REQUIRE Configuration parameters
\ENSURE Current navigation state estimate $\boldsymbol{X}$
\STATE Initialize system, IMU buffer, and depth buffer
\IF{Initial-alignment command received}
    \STATE Perform initial alignment in dynamic frame
\ENDIF
\WHILE{System is running}

    \IF{$\boldsymbol{Z}_{\text{piUSBL}}$ available}
        \IF{Time-delay compensation enabled}
            \STATE Measure $\Delta t$ and set $t_1 \leftarrow t_4-\Delta t$
            \STATE Compute IMU interval $\Delta t_{\mathrm{IMU}}$ and epoch mismatch $e \leftarrow |t_4-t_1|$
            \IF{$e < 0.1\,\Delta t_{\mathrm{IMU}}$}
                \STATE Obtain $\tilde{Z}_{\text{Depth}}(t_4)$ from depth buffer
                \STATE Joint update using $\{\boldsymbol{Z}_{\text{piUSBL}}, \tilde{Z}_{\text{Depth}}(t_4)\}$ at $t_4$
            \ELSE
                \STATE Retrieve buffered state $\boldsymbol{X}(t_1)$ from buffer
                \STATE Obtain $\tilde{Z}_{\text{Depth}}(t_1)$ from depth buffer
                \STATE \textbf{Start FR thread:} joint update at $t_1$ and re-propagate to $t_4$ using buffered IMU
                \STATE Replace current estimate by the FR result at $t_4$
            \ENDIF
        \ELSE
            \STATE Obtain $\tilde{Z}_{\text{Depth}}(t_4)$ from depth buffer
            \STATE Joint update using $\{\boldsymbol{Z}_{\text{piUSBL}}, \tilde{Z}_{\text{Depth}}(t_4)\}$ at $t_4$
        \ENDIF
        \STATE Publish and log results
    \ENDIF

    \IF{$Z_{\text{Depth}}$ available}
        \STATE Append $Z_{\text{Depth}}$ to depth buffer
        \STATE Depth-only update at its timestamp
    \ENDIF

    \IF{$\boldsymbol{Z}_{\text{IMU}}$ available}
        \STATE Append $\boldsymbol{Z}_{\text{IMU}}$ to IMU buffer
        \STATE Propagate (mechanize) $\boldsymbol{X}$ using $\boldsymbol{Z}_{\text{IMU}}$
    \ENDIF
\ENDWHILE
\end{algorithmic}
\end{algorithm}
}

\bibliographystyle{IEEEtran}
\bibliography{ref}

@article{paullAUVNavigationLocalization2014,
  title = {AUV Navigation and Localization: A Review},
  author = {Paull, Liam and Saeedi, Sajad and Seto, Mae and Li, Howard},
  year = {2014},
  journal = {IEEE Journal of Oceanic Engineering},
  volume = {39},
  number = {1},
  pages = {131--149},
  issn = {0364-9059},
  doi = {10.1109/JOE.2013.2278891},
  lccn = {2}
}

@article{liImprovedESObasedLineofSight2025,
  title = {An Improved ESO-Based Line-of-Sight Guidance Law for Path Following of Underactuated Autonomous Underwater Helicopter with Nonlinear Tracking Differentiator and Anti-Saturation Controller},
  author = {Li, Haoda and Liu, Zichen and Huang, Jin and An, Xinyu and Chen, Ying},
  year = {2025},
  journal = {Ocean Engineering},
  volume = {322},
  pages = {120456},
  issn = {0029-8018},
  doi = {10.1016/j.oceaneng.2025.120456},
  urldate = {2025-02-03},
  lccn = {2}
}

@article{zhangAutonomousUnderwaterVehicle2023,
  title = {Autonomous Underwater Vehicle Navigation: A Review},
  author = {Zhang, Bingbing and Ji, Daxiong and Liu, Shuo and Zhu, Xinke and Xu, Wen},
  year = {2023},
  journal = {Ocean Engineering},
  volume = {273},
  pages = {113861},
  issn = {0029-8018},
  doi = {10.1016/j.oceaneng.2023.113861},
  urldate = {2023-02-27},
  lccn = {2}
}

@article{chenReviewAUVUnderwater2015,
  title = {Review of AUV Underwater Terrain Matching Navigation},
  author = {Chen, Pengyun and Li, Ye and Su, Yumin and Chen, Xiaolong and Jiang, Yanqing},
  year = {2015},
  journal = {Journal of Navigation},
  volume = {68},
  number = {6},
  pages = {1155--1172},
  issn = {0373-4633},
  doi = {10.1017/S0373463315000429},
  urldate = {2024-05-31},
  lccn = {3}
}

@article{millerAutonomousUnderwaterVehicle2010,
  title = {Autonomous Underwater Vehicle Navigation},
  author = {Miller, Paul A. and Farrell, Jay A. and Zhao, Yuanyuan and Djapic, Vladimir},
  year = {2010},
  journal = {IEEE Journal of Oceanic Engineering},
  volume = {35},
  number = {3},
  pages = {663--678},
  issn = {0364-9059},
  doi = {10.1109/JOE.2010.2052691},
  urldate = {2024-05-30},
  lccn = {2}
}

@article{maurelliAUVLocalisationReview2022,
  title = {AUV Localisation: A Review of Passive and Active Techniques},
  author = {Maurelli, Francesco and Krupi{\'n}ski, Szymon and Xiang, Xianbo and Petillot, Yvan},
  year = {2022},
  journal = {International Journal of Intelligent Robotics and Applications},
  volume = {6},
  number = {2},
  pages = {246--269},
  issn = {2366-5971},
  doi = {10.1007/s41315-021-00215-x},
  urldate = {2024-05-31}
}

@inproceedings{zhaoReviewUnderwaterMultisource2023,
  title = {A Review of Underwater Multi-Source Positioning and Navigation Technology},
  booktitle = {Lecture Notes in Electrical Engineering},
  author = {Zhao, Wanlong and Qi, Shuaijie and Liu, Ruitong and Zhang, Guoyao and Liu, Gongliang},
  editor = {Yan, Liang and Duan, Haibin and Deng, Yimin},
  year = {2023},
  series = {Lecture Notes in Electrical Engineering},
  pages = {5466--5479},
  publisher = {Springer Nature},
  address = {Singapore},
  doi = {10.1007/978-981-19-6613-2_528},
  isbn = {978-981-19-6613-2}
}

@inproceedings{vickeryAcousticPositioningSystems1998,
  title = {Acoustic Positioning Systems. New Concepts-the Future},
  booktitle = {Proceedings of the 1998 Workshop on Autonomous Underwater Vehicles (Cat. No.98CH36290)},
  author = {Vickery, K.},
  year = {1998},
  pages = {103--110},
  doi = {10.1109/AUV.1998.744445},
  urldate = {2025-08-05}
}

@inproceedings{rypkemaOnewayTraveltimeInverted2017,
  title = {One-Way Travel-Time Inverted Ultra-Short Baseline Localization for Low-Cost Autonomous Underwater Vehicles},
  booktitle = {2017 IEEE International Conference on Robotics and Automation (ICRA)},
  author = {Rypkema, Nicholas R. and Fischell, Erin M. and Schmidt, Henrik},
  year = {2017},
  pages = {4920--4926},
  publisher = {IEEE},
  doi = {10.1109/ICRA.2017.7989570}
}

@article{guoRobustAttitudeEstimation2023,
  title = {A Robust Attitude Estimation Algorithm for Seabed Inverted Ultra-Short Baseline},
  author = {Guo, Haoran and Qian, Zhiwen and Wang, Xiaojian and Sun, Wanzhong and Jie, Li and Zhai, Jingsheng},
  year = {2023},
  journal = {Ocean Engineering},
  volume = {280},
  pages = {114534},
  issn = {0029-8018},
  doi = {10.1016/j.oceaneng.2023.114534},
  urldate = {2025-08-05},
  lccn = {2}
}

@inproceedings{rypkemaPassiveInvertedUltrashort2019,
  title = {Passive Inverted Ultra-Short Baseline (piUSBL) Localization: An Experimental Evaluation of Accuracy},
  booktitle = {2019 IEEE/RSJ International Conference on Intelligent Robots and Systems (IROS)},
  author = {Rypkema, Nicholas R. and Schmidt, Henrik},
  year = {2019},
  pages = {7197--7204},
  publisher = {IEEE},
  doi = {10.1109/IROS40897.2019.8967800}
}

@inproceedings{ridaoUSBLDVLNavigation2011,
  title = {USBL/DVL Navigation through Delayed Position Fixes},
  booktitle = {2011 IEEE International Conference on Robotics and Automation},
  author = {Ridao, Pere and Ribas, David and Hernandez, Emili and Rusu, Alex},
  year = {2011},
  pages = {2344--2349},
  doi = {10.1109/ICRA.2011.5980110},
  urldate = {2025-07-20}
}

@article{fossenFeedbackErrorstateKalman2023,
  title = {Feedback Error-State Kalman Filter with Time-Delay Compensation for Hydroacoustic-Aided Inertial Navigation of Underwater Vehicles},
  author = {Fossen, Thor I.},
  year = {2023},
  journal = {Control Engineering Practice},
  volume = {138},
  pages = {105603},
  issn = {0967-0661},
  doi = {10.1016/j.conengprac.2023.105603},
  urldate = {2023-07-16},
  lccn = {2}
}

@article{xiaMixtureDistributionBasedRobust2021,
  title = {A Mixture Distribution-Based Robust SINS/USBL Integration Navigation With Time-Varying Delays},
  author = {Xia, Maodong and Zhang, Tao and Zhang, Liang and Yang, Bo and Shi, Yang},
  year = {2021},
  journal = {Measurement Science and Technology},
  volume = {32},
  number = {1},
  pages = {015903},
  issn = {0957-0233},
  doi = {10.1109/TIM.2024.3381293},
  urldate = {2025-07-20},
  lccn = {2}
}

@article{wangAdaptiveUnscentedKalman2025a,
  title = {Adaptive Unscented Kalman Filtering and Signal-Sending Time Estimation for Calibration of SINS/USBL Navigation System},
  author = {Wang, Guolin and Hu, Shuibo and Liu, Mengzhuo and Cui, Jiankuo and Peng, Zheng and Cui, Jun-Hong},
  year = {2025},
  journal = {IEEE Transactions on Instrumentation and Measurement},
  volume = {74},
  pages = {1--12},
  issn = {0018-9456},
  doi = {10.1109/TIM.2025.3571127},
  urldate = {2025-07-20},
  lccn = {2}
}

@article{comelliniIncorporatingDelayedMultirate2020,
  title = {Incorporating Delayed and Multirate Measurements in Navigation Filter for Autonomous Space Rendezvous},
  author = {Comellini, Anthea and Casu, Davide and Zenou, Emmanuel and Dubanchet, Vincent and Espinosa, Christine},
  year = {2020},
  journal = {Journal of Guidance, Control, and Dynamics},
  volume = {43},
  number = {6},
  pages = {1164--1172},
  issn = {1533-3884},
  doi = {10.2514/1.G005034},
  urldate = {2025-07-21},
  copyright = {Copyright {\copyright} 2020 by the American Institute of Aeronautics and Astronautics, Inc. All rights reserved. All requests for copying and permission to reprint should be submitted to CCC at www.copyright.com; employ the eISSN 1533-3884 to initiate your request. See also AIAA Rights and Permissions www.aiaa.org/randp.}
}

@article{xuMaximumCorrentropyDelay2021a,
  title = {Maximum Correntropy Delay Kalman Filter for SINS/USBL Integrated Navigation},
  author = {Xu, Bo and Wang, Xiaoyu and Zhang, Jiao and Razzaqi, Asghar A. and Xu, Bo and Wang, Xiaoyu and Zhang, Jiao and Razzaqi, Asghar A. and Xu, Bo and Wang, Xiaoyu and Zhang, Jiao and Razzaqi, Asghar A. and Xu, Bo and Wang, Xiaoyu and Zhang, Jiao and Razzaqi, Asghar A.},
  year = {2021},
  journal = {ISA Transactions},
  volume = {117},
  pages = {274--287},
  issn = {0019-0578},
  doi = {10.1016/j.isatra.2021.01.055},
  urldate = {2025-07-20},
  lccn = {2}
}

@inproceedings{ribasDelayedStateInformation2012,
  title = {Delayed State Information Filter for USBL-Aided AUV Navigation},
  booktitle = {2012 IEEE International Conference on Robotics and Automation},
  author = {Ribas, David and Ridao, Pere and Mallios, Angelos and Palomeras, Narcis},
  year = {2012},
  pages = {4898--4903},
  doi = {10.1109/ICRA.2012.6224989},
  urldate = {2025-07-20}
}

@article{liuRobustTightlySINS2023a,
  title = {A Robust Tightly SINS/USBL Based AUV Localization Method Aided by Dual Transponder},
  author = {Liu, Shede and Zhang, Tao and Zhang, Jiayu and Xia, Maodong},
  year = {2023},
  journal = {Measurement},
  volume = {223},
  pages = {113725},
  issn = {0263-2241},
  doi = {10.1016/j.measurement.2023.113725},
  urldate = {2025-07-20},
  lccn = {2}
}

@article{xiaoAcousticCommunicationTime2017,
  title = {An Acoustic Communication Time Delays Compensation Approach for Master--Slave AUV Cooperative Navigation},
  author = {Xiao, Guangdi and Wang, Bo and Deng, Zhihong and Fu, Mengyin and Ling, Yun},
  year = {2017},
  journal = {IEEE Sensors Journal},
  volume = {17},
  number = {2},
  pages = {504--513},
  issn = {1530-437X},
  doi = {10.1109/JSEN.2016.2631478},
  urldate = {2025-07-20},
  lccn = {3}
}

@article{xuNovelRobustFilter2021a,
  title = {A Novel Robust Filter for Outliers and Time-Varying Delay on an SINS/USBL Integrated Navigation Model},
  author = {Xu, Bo and Zhang, Jiao and Razzaqi, Asghar A},
  year = {2021},
  journal = {Measurement Science and Technology},
  volume = {32},
  number = {1},
  pages = {015903},
  issn = {0957-0233},
  doi = {10.1088/1361-6501/abaae9},
  urldate = {2025-07-20},
  lccn = {3}
}

@article{wangPassiveInvertedUltraShort2022,
  title = {Passive Inverted Ultra-Short Baseline Positioning for a Disc-Shaped Autonomous Underwater Vehicle: Design and Field Experiments},
  author = {Wang, Yingqiang and Hu, Ruoyu and Huang, S. H. and Wang, Zhikun and Du, Peizhou and Yang, Wencheng and Chen, Ying},
  year = {2022},
  journal = {IEEE Robotics and Automation Letters},
  volume = {7},
  number = {3},
  pages = {6942--6949},
  issn = {2377-3766},
  doi = {10.1109/LRA.2022.3178494},
  urldate = {2024-05-07},
  lccn = {2}
}

@article{wangDesignExperimentalResults2022,
  title = {Design and Experimental Results of Passive iUSBL for Small AUV Navigation},
  author = {Wang, Yingqiang and Huang, S. H. and Wang, Zhikun and Hu, Ruoyu and Feng, Mingyue and Du, Peizhou and Yang, Wencheng and Chen, Ying},
  year = {2022},
  journal = {Ocean Engineering},
  volume = {248},
  pages = {110812},
  issn = {0029-8018},
  doi = {10.1016/j.oceaneng.2022.110812},
  urldate = {2022-02-20},
  lccn = {3.795}
}

@inproceedings{wangInformationFusionAlgorithm2023a,
  title = {An Information Fusion Algorithm for INS/DVL/USBL Integrated Navigation System Based on Delay State Unscented Kalman Filter},
  booktitle = {2023 IEEE International Conference on Mechatronics and Automation (ICMA)},
  author = {Wang, Xi and Wei, Yanhui and Shao, Hong},
  year = {2023},
  volume = {32},
  pages = {1161--1166},
  doi = {10.1109/ICMA57826.2023.10216046},
  urldate = {2025-07-20}
}

@article{xuRobustIterativeAlgorithm2025,
  title = {A Robust Iterative Algorithm for SINS/USBL Integrated Navigation Based on Dual Hydrophone Differential Model},
  author = {Xu, Bo and Zhu, Haibin and Guo, Yu},
  year = {2025},
  journal = {Measurement},
  volume = {242},
  pages = {115854},
  issn = {0263-2241},
  doi = {10.1016/j.measurement.2024.115854},
  urldate = {2025-07-20},
  lccn = {2}
}

@book{grovesPrinciplesGNSSInertial2013,
  title = {Principles of GNSS, Inertial, and Multisensor Integrated Navigation Systems, Second Edition},
  author = {Groves, Paul D.},
  year = {2013},
  publisher = {Artech House},
  isbn = {978-1-60807-005-3}
}

@article{wangAdaptiveUnscentedKalman2025,
  title = {Adaptive Unscented Kalman Filtering and Signal-Sending Time Estimation for Calibration of SINS/USBL Navigation System},
  author = {Wang, Guolin and Hu, Shuibo and Liu, Mengzhuo and Cui, Jiankuo and Peng, Zheng and Cui, Jun-Hong},
  year = {2025},
  journal = {IEEE Transactions on Instrumentation and Measurement},
  volume = {74},
  pages = {1--12},
  issn = {0018-9456},
  doi = {10.1109/TIM.2025.3571127},
  urldate = {2025-06-16},
  lccn = {2}
}

@article{niuDevelopmentEvaluationGNSS2015,
  title = {Development and Evaluation of GNSS/INS Data Processing Software for Position and Orientation Systems},
  author = {Niu, X. and Zhang, Q. and Gong, L. and Liu, C. and Zhang, H. and Shi, C. and Wang, J. and Coleman, M.},
  year = {2015},
  journal = {Survey Review},
  volume = {47},
  number = {341},
  pages = {87--98},
  issn = {0039-6265},
  doi = {10.1179/1752270614Y.0000000099},
  urldate = {2023-12-21},
  lccn = {4}
}

@book{tittertonStrapdownInertialNavigation2004,
  title = {Strapdown Inertial Navigation Technology},
  author = {Titterton, David and Weston, John},
  year = {2004},
  publisher = {IET Digital Library},
  doi = {10.1049/PBRA017E},
  urldate = {2024-05-22},
  isbn = {978-1-84919-093-0}
}

@article{wangRobustFilterMethod2023,
  title = {Robust Filter Method for SINS/DVL/USBL Tight Integrated Navigation System},
  author = {Wang, Di and Wang, Bing and Huang, Haoqian and Yao, Yiqing and Xu, Xiang},
  year = {2023},
  journal = {IEEE Sensors Journal},
  volume = {23},
  doi = {10.1109/JSEN.2023.3264755},
  lccn = {2}
}

@article{maStateArtKey2025,
  title = {The State of the Art in Key Technologies for Autonomous Underwater Vehicles: A Review},
  author = {Ma, Dong and Li, Ye and Ma, Teng and Pascoal, Ant{\'o}nio M.},
  year = {2025},
  journal = {Engineering},
  issn = {2095-8099},
  doi = {10.1016/j.eng.2025.08.002},
  urldate = {2025-08-12},
  lccn = {2}
}

@phdthesis{rypkemaUnderwaterOutSight2019,
  type = {Thesis},
  title = {Underwater \& out of Sight : Towards Ubiquity in Underwater Robotics},
  author = {Rypkema, Nicholas Rahardiyan},
  year = {2019},
  copyright = {MIT theses are protected by copyright. They may be viewed, downloaded, or printed from this source but further reproduction or distribution in any format is prohibited without written permission.},
  school = {Massachusetts Institute of Technology}
}

@article{duNovelLieGroup2022,
  title = {A Novel Lie Group Framework-Based Student's {\emph{t}} Robust Filter and Its Application to INS/DVL Tightly Integrated Navigation},
  author = {Du, Siyuan and Zhu, Fengchi and Wang, Zhao and Huang, Yulong and Zhang, Yonggang and Du, Siyuan and Zhu, Fengchi and Wang, Zhao and Huang, Yulong and Zhang, Yonggang and Du, Siyuan and Zhu, Fengchi and Wang, Zhao and Huang, Yulong and Zhang, Yonggang},
  year = 2022,
  journal = {IEEE Transactions on Instrumentation and Measurement},
  volume = {71},
  pages = {1--27},
  issn = {0018-9456},
  doi = {10.1109/TIM.2024.3379400},
  urldate = {2025-03-23},
  lccn = {2}
}

@article{zhangNovelINSUSBL2022,
  title = {A Novel INS/USBL Integrated Navigation Scheme via Factor Graph Optimization},
  author = {Zhang, Liang and Hsu, Li-Ta and Zhang, Tao},
  year = {2022},
  journal = {IEEE Transactions on Vehicular Technology},
  volume = {71},
  number = {9},
  pages = {9239--9249},
  issn = {0018-9545},
  doi = {10.1109/TVT.2022.3177739},
  urldate = {2025-10-10},
  lccn = {2}
}

@article{huangGNSSaidedInstallationError2025,
  title = {GNSS-Aided Installation Error Compensation for DVL/INS Integrated Navigation System Using Error-State Kalman Filter},
  author = {Huang, Jin and Li, Haoda and Liu, Zichen and Wang, Zhikun and Wang, Yingqiang and Chen, Ying},
  year = {2025},
  journal = {Measurement},
  volume = {242},
  pages = {116224},
  issn = {0263-2241},
  doi = {10.1016/j.measurement.2024.116224},
  urldate = {2025-01-10},
  lccn = {2}
}

@article{yuUnderwaterLocalizationAUVs2023a,
  title = {Underwater Localization of AUVs in Motion Using Two-Way Travel Time Measurements With Unknown Sound Velocity},
  author = {Yu, Xiang and Qin, Hong-De and Zhu, Zhong-Ben},
  year = {2023},
  journal = {IEEE Transactions on Vehicular Technology},
  volume = {72},
  number = {9},
  pages = {11358--11373},
  issn = {0018-9545},
  doi = {10.1109/TVT.2023.3270931},
  urldate = {2025-10-10},
  lccn = {2}
}

@article{zhaoUnsynchronizedUnderwaterLocalization2024,
  title = {Unsynchronized Underwater Localization With Isogradient Sound Speed Profile and Anchor Location Uncertainties},
  author = {Zhao, Haiyan and Gong, Zijun and Yan, Jing and Li, Cheng and Guan, Xinping},
  year = {2024},
  journal = {IEEE Transactions on Vehicular Technology},
  volume = {73},
  number = {6},
  pages = {8864--8877},
  issn = {0018-9545},
  doi = {10.1109/TVT.2024.3360252},
  urldate = {2025-10-14},
  lccn = {2}
}

@article{xiaMixtureDistributionBasedRobust2024,
  title = {A Mixture Distribution-Based Robust SINS/USBL Integration Navigation With Time-Varying Delays},
  author = {Xia, Maodong and Zhang, Tao and Zhang, Liang and Yang, Bo and Shi, Yang},
  year = 2024,
  journal = {IEEE Transactions on Instrumentation and Measurement},
  volume = {73},
  pages = {1--10},
  issn = {0018-9456},
  doi = {10.1109/TIM.2024.3381293},
  urldate = {2024-04-01},
  lccn = {2}
}

@article{zhangNovelUnderwaterSINS2025,
  title = {A Novel Underwater SINS/USBL Dual-Responder Integrated Navigation Algorithm Considering Motion Effects},
  author = {Zhang, Shuaishuai and Zhang, Tao and Li, Shengxin and Zhang, Liang and Shi, Yang},
  year = 2025,
  journal = {IEEE Transactions on Instrumentation and Measurement},
  volume = {74},
  pages = {1--13},
  issn = {0018-9456},
  doi = {10.1109/TIM.2025.3580832},
  urldate = {2025-07-08},
  lccn = {2}
}

@misc{huangRaspi2USBLOpensourceRaspberry2025,
  title = {Raspi\$\textasciicircum 2\$USBL: An Open-Source Raspberry Pi-Based Passive Inverted Ultra-Short Baseline Positioning System for Underwater Robotics},
  author = {Huang, Jin and Wang, Yingqiang and Chen, Ying},
  year = 2025,
  number = {arXiv:2511.06998},
  eprint = {2511.06998},
  primaryclass = {cs},
  publisher = {arXiv},
  doi = {10.48550/arXiv.2511.06998},
  urldate = {2025-11-11},
  archiveprefix = {arXiv}
}

@inproceedings{leeStateParameterEstimation2017,
  title = {State and Parameter Estimation Using Measurements with Unknown Time Delay},
  booktitle = {2017 IEEE Conference on Control Technology and Applications (CCTA)},
  author = {Lee, Kyuman and Johnson, Eric N.},
  year = 2017,
  pages = {1402--1407},
  doi = {10.1109/CCTA.2017.8062655},
  urldate = {2026-01-11}
}

@article{niuKFGINSOpensourcedSoftware2025,
  title = {KF-GINS: An Open-Sourced Software for GNSS/INS Integrated Navigation},
  author = {Niu, Xiaoji and Wang, Liqiang and Chen, Qijin and Tang, Hailiang and Zhang, Quan and Zhang, Tisheng},
  year = 2025,
  journal = {GPS Solutions},
  volume = {29},
  number = {4},
  pages = {202},
  issn = {1080-5370},
  doi = {10.1007/s10291-025-01967-w},
  urldate = {2025-10-02},
  lccn = {1}
}

\end{document}